\documentclass[12pt]{article}

\usepackage[paper=portrait,pagesize]{typearea}

\usepackage{bbm}

\usepackage{setspace}
\usepackage{helvet}

\usepackage{footmisc} 

\usepackage[utf8]{inputenc}
\usepackage[authordate, backend=biber]{biblatex-chicago}
\DeclareBibliographyDriver{data}{%
  \printnames{author}%
  \setunit{\labelnamepunct}%
  \printfield{title}%
  \setunit{\addcomma\space}%
  \printfield{year}%
  \finentry
}

\usepackage[table]{xcolor} 
\usepackage{booktabs, makecell}

\usepackage[labelfont=bf, font={footnotesize}]{caption}[2005/07/16]
\usepackage[font={footnotesize},labelformat=simple]{subcaption}

\usepackage{dsfont}

\usepackage{algorithm}
\usepackage{algpseudocode}
\makeatletter
\renewcommand{\ALG@name}{Procedure}
\makeatother

\usepackage{enumitem}

\usepackage{amsmath}
\usepackage{amsthm}

\usepackage{pdflscape} 

\usepackage{array}

\usepackage{float}
\usepackage{graphicx} 
\usepackage{longtable} 
\usepackage{dcolumn}

\usepackage{subfiles}

\usepackage{hyperref}

\title{Exploring the Potential Role of Generative AI in the TRAPD Procedure for Survey Translation\thanks{We gratefully acknowledge FORMAS (2016-00228, PI: Ellen Lust) and The Swedish Research Council (2016-01687, PI: Ellen Lust, and E0003801, PI: Pam Fredman) for financial support; the Mixed Methods for Analyzing Communication (MiMac) Conference for early encouragement; the Governance and Local Development Institute (GLD) for insights and feedback; and the World Association of Public Opinion Research (WAPOR) 2024 conference for feedback.}}
\author{Erica Ann Metheney\thanks{Researcher/Statistician/Head of Data, Governance and Local Development Institute, University of Gothenburg}\\
Lauren Yehle\thanks{Associate Researcher, University of Gothenburg}}

\date{}

\usepackage{appendix} 
\usepackage{threeparttable}

\begin{document}

\maketitle
\thispagestyle{empty}
\newpage


\thispagestyle{empty}

\begin{abstract} 
This paper explores and assesses in what ways generative AI can assist in translating survey instruments. Writing effective survey questions is a challenging and complex task, made even more difficult for surveys that will be translated and deployed in multiple linguistic and cultural settings. Translation errors can be detrimental, with known errors rendering data unusable for its intended purpose and undetected errors leading to incorrect conclusions. A growing number of institutions face this problem as surveys deployed by private and academic organizations globalize, and the success of their current efforts depends heavily on researchers’ and translators’ expertise and the amount of time each party has to contribute to the task. Thus, multilinguistic and multicultural surveys produced by teams with limited expertise, budgets, or time are at significant risk for translation-based errors in their data. We implement a zero-shot prompt experiment using ChatGPT to explore generative AI’s ability to identify features of questions that might be difficult to translate to a linguistic audience other than the source language. We find that ChatGPT can provide meaningful feedback on translation issues, including common source survey language, inconsistent conceptualization, sensitivity and formality issues, and nonexistent concepts. In addition, we provide detailed information on the practicality of the approach, including accessing the necessary software, associated costs, and computational run times. Lastly, based on our findings, we propose avenues for future research that integrate AI into survey translation practices.

\end{abstract}

\newpage

\setcounter{page}{1}

\section{Introduction}\label{sec:introduction}

Structured questionnaires with set wording and order are a primary instrument of survey research and closely tied to quantitative analysis \autocite{cheung_structured_2014}. Good survey questions are both reliable and valid measures of \textit{something else}; this is to say researchers are not interested in the answer to the question for its own sake but to what extent the answer can inform a researcher's theoretical or empirical interests \autocite{fowler_improving_1995}. Relaying the correct type of answer to gather meaningful results becomes more complex as researchers translate questions in an attempt to conduct 3M surveys (multi-cultural, lingual, and/or national) \autocite{harkness_survey_2004}. 

This paper explores the possibility of using generative AI to assist in preparing survey questions for translators via a computational experiment. The release of ChatGPT, developed by OpenAI, as a free, widely accessible large language model (LLM) chat bot has been met with celebration and caution for its capacity to produce highly realistic texts in regards to logic, persuasiveness, and originality \autocite{liu2023summary}. Peer reviewed, published papers and ethical standards struggle to keep pace with the development and potential capabilities of ChatGPT--3.5 released November 2022, its successor GPT--4 released March 2023, and the most recent model GPT--4 omni (GPT--4o) released May 2024. Of working and published studies regarding generative AI, most studies address the application to fields like education, legal, and computer science rather than research applications for social science \autocites {liu2023summary,wu2023brief}. Some work pertains to translation and the comparatively better translations than other free online tools \autocite{lund2023chatting}, but no work to the authors' knowledge has been conducted regarding preparing survey questions for multi linguistic and multi cultural settings.  

Liu et al (2023) suggest that constructive human--LLM collaboration would entail humans providing the expertise, creativity, and decision-making abilities while the machine provides automation, scalability, and computing power. We see this suggestion applied to the preparation of a survey for translation with questions that better relay the researcher's intent. We do not argue or believe that generative AI can replace survey best practices but rather recognize that question design, pre--testing techniques, and translation can still fall prey to human error, particularly under time, funding, or expertise constraints. Current best practices in survey design follow the total survey error paradigm  which calls for balancing quality with the reality of budgetary and time constraints \autocite{biemer2010total}. We believe that generative AI has a strong role to play in this paradigm. 

An AI identifying problems that would have been overlooked lessens the high costs affiliated with discarding responses after conducting the survey. Good questionnaire design in the source language is a vital precondition for translation, but even if a perfect question is developed in the source language, researchers often lose control or understanding of their research instruments upon translation \autocite{behr2016translation}. Identifying translation problems early in the process so primary researchers can decide to adapt, modify, or proceed with the question is a vital aspect of ``advance translation" \autocite{behr2016translation}. When translators do not understand the research objective, they often make a personal decision that does not align with the researcher's research goals or understanding of the target linguistic audience \autocite{harkness_questionnaires_1998}. The LLM can assist primary researchers in preparing a version of the survey stripped of cultural significant and linguistically embedded vocabulary for translators to write conceptually equivalent questions \autocite{erkut2010developing}. This higher level of translation control allows researchers to better understand confounding effects after collecting data \autocite{erkut2010developing}. With time and integration to survey translation, LLM tools may likely be faster and more efficient than if the principle investigators and team of translators had developed the survey without it.

We conduct a computational experiment to explore generative AI's ability to assist in preparing survey questions for translation.  Though this may seem like an elementary use of generative AI and LLMs, the volume of information that these tools can provide lessens the burden to any researcher or research team to oversee the numerous linguistic and cultural issues with surveys in translation. These problems are costly, may make it difficult to compare between cultures as valid instruments, and at the worst may render results incomprehensible entirely.  These problems are  exacerbated and  likely to occur as surveys conducted by private organizations like Pew and Gallup and academic organizations like World Values Survey, Afro-barometer, and European social survey globalize to dozens if not hundreds of countries and linguistic groups \autocite{smith_sage_2016}. 

This study rests on the idea that the AI may identify problems that a researcher may be too tired, inexperienced, or under resourced to find. An illuminating example of this would be asking a Chinese person for the size of his or her household in a large, cross cultural survey conducted in many countries as part of the basic demographic battery of questions. While most people and researchers are aware of the one child policy present from 1980 to 2016 in the People's Republic of China, it would be easy to forget and misinterpret why Chinese participants are (1) hesitant to answer and/or (2) officially reporting households of three especially when comparing these results to one of many other countries. This type of problem may get lost in the sea of translation problems like syntax, word use, formality, and utilizing characters instead of an alphabet. Thus, AI identifying this problem lessens the load for researchers beforehand and renders the analysis more meaningful after data collection. In additional to exploring the capacity for generative AI's capacity to help in questionnaire translation, we provide a guide for non-computer scientists to best utilize accessible tools like ChatGPT--3.5 and GPT--4 in terms of time, money, and capability. 

This paper is laid out as follows: in Section 2, we present ongoing research about generative AI and best practices for 3M survey questions. In Section 3, we present our research questions. Section 4 outlines the methodology outlining the experimental design, data collection procedure, and analysis plan. We follow this with the results in Section 5. Finally, Section 6 discusses the results and Section 7 concludes. 

\section{Previous research}\label{litreview}

Robust questions in the source language are a necessary pre-condition before translating and conducting the survey in secondary cultures/or languages. A vast literature exists regarding how to design source questions \autocite{converse_survey_1986, fowler_improving_1995, willis2016questionnaire} that we do not address here. Instead, we focus on the ways survey development becomes even more complex as researchers study across country, linguistic, and cultural borders \autocite{behr2016translation}.

\subsection{Questionnaires in Translation}
 Unlike literal translation, standard practice for survey translation attempts to produce \textit{equivalent} survey instruments to minimize error and resultant data contamination so that researchers have valid, reliable, and comparable responses across linguistic and cultural barriers \autocites{weeks2007issues, behling2000translating, harkness_questionnaires_1998}.  However, primary researchers may not be experts in all the countries and linguistic groups in which they conduct research; the primary problem is thus bridging the researcher's theoretical interests with equivalent concepts across linguistic and cultural differences. 
 
 There are at least three practical problems associated with equivalency: semantic in languages, conceptual in cultures, and normative behaviors in societies \autocite{behling2000translating}. Some equivalency issues can be determined for the survey as a whole, such as formality, register, and regional dialects, but other problems are more embedded in specific questions. In preparing the questions for translation, \cite{weeks2007issues} suggest a number of solutions to minimize error affiliated with equivalency problems like simple sentences shorter than 16 words, repeating nouns frequently, active not passive voice, avoiding metaphors and colloquialisms, specific rather than generic terms, using a range of synonyms, amongst others. However, these general recommendations fail to provide specific nuance for every question.

With all these potential threats to equivalency, the state-of-the-art procedure to produce high quality surveys is now called the Translation, Review, Adjudication, Pre-Test, and Documentation (TRAPD) method. The method convenes researchers and translators to collaboratively produce the translation in 5 steps \autocite{UMichGuidelines}. In the Translation phase, two or more translators produce independent translations of the survey instrument. There are no set instructions on how to produce this first set of translations.  That being said, \cite{harkness_survey_2004} highlights that translators need solid support materials including example texts, information regarding the target audience, administration mode, and clear insight on the conceptual intent to produce quality translations. In the Review phase, the translators and researchers come together to agree on a final version. During the next phase (A), the adjudicator determines if the translation is ready move on to the pre-testing phase. Pre-testing most often includes pilot testing to determine if any additional translation modifications are necessary. Throughout the process, the team documents their activities (step D). Though this process still compensates for many potential translation problems, space still remains costly error. Also, this ideal process might be shortened under time and money constraints increasingly the likelihood of error.

\subsection{Generative AI Overview}\label{subsec:genAI}
Generative AI is a type of deep learning model that can create new content similar to the training content \autocite{google_cloud_tech_introduction_2023}. This new iteration of AI was made possible using transformers (the T in ChatGPT) to encode and decode the neural networks necessary for machine learning. This differs from discriminative AI used primarily to classify, predict, and cluster information by learning relationships between features of data points but does not create new data. Both type of AI are statistical predictive models and are trained by structured (human labeled) and unstructured data. A discriminative AI is well equipped to produce numbers, class, and probabilities e.g., identify an image of a dog, while generative AI models are better equipped to produce novel natural language, image, and audio e.g., create a dog image. 

ChatGPT's models are examples of a new type of AI system called the ``foundation model,” also known as the “general purpose AI” or “GPAI”  because these chatbots can handle a wide range of tasks \autocite{jones_explainer_2023}. ChatGPT and similar tools are a large language model (LLM), which generates convincing sentences by mimicking the statistical patterns of language from a huge database of text collated from the internet \autocite{stokel2023chatgpt}. ChatGPT has over 175 billion parameters trained by a diverse dataset of naturally used text obtained from different sources like web pages, books, research articles, and social chatter \autocite{dwivedi2023so}. Foundation model AIs lack a designated purpose, so output may be too vague, ambiguous, lengthy, or brief which may be caused by insufficient, noisy, or uncontextualized training data \autocite{jones_explainer_2023}. A way to compensate for these responses may be using prompts, i.e. instructions given to an LLM to enforce rules, automate processes, and ensure specific qualities and quantities of output \autocite{white_prompt_2023}.

Peer reviewed research is relatively limited regarding the uses and ethical application of Generative AI due to the recent release of broadly used, easily accessible LLMs versus the timeline of peer reviewed research. That aside, preliminary research indicates that ChatGPT outperforms various state of the art zero shot learning large language models in most tasks especially when the AI is given more fine-tuned instructions \autocite{liu2023summary}. Liu (2023) suggests using tools like ChatGPT for semantic fields of communication where chat--gpt can act as ``an intelligent consulting assistant" in identifying the semantic importance of words in messages. Additionally, findings by \cite{hartmann2023political} indicate ChatGPT attempts to maintain a neutral position but when \textit{pressed} has a left-libertarian ideology that is pro-environment. This may lead to potential biases when asking for opinions and is why we refrain from pretesting survey questions on the AI. \cite{borji2023categorical} categorizes the types of errors ChatGPT makes as most affiliated with spatial, temporal and spatial reasoning; other problems can be common sensical like humor and sarcasm or psychological like attributing false beliefs to others. Borji (2023) highlights that ChatGPT tends to be ``excessively comprehensive and verbose, approaching a topic from multiple angles;" whether this is excessive to a fault depends on the task. Also, Borji (2023) notes that AI is especially good at working in multiple languages like English, Spanish, French, and German. Many of these problems may improve as the tools are updated with new training material. Though this study uses ChatGPT and GPT--4, other and future LLMs should be able to handle similar tasks to varying qualitative degrees.  

\section{Research Questions}

Based on the total survey error paradigm, we expect generative AI tools, specifically ChatGPT, to supplement survey questionnaire development. We do not expect AI to replace any specific survey development best practices but rather reveal problems that humans may miss, particularly when facing logistical constraints. We expect the AI to be especially well equipped to assist with preparing annotated versions of surveys to give translators. For languages with a large online presence, we expect the AI tools to indicate which concepts may be culturally, linguistically, or normatively difficult to create an equivalent instrument. This would provide vital information to the primary researcher, so that the translation is more under her control than if the translators make the decision themselves. The AI tool may also have access to more information about culturally sensitive and nonexistent topics, so the researcher may be warned before pre-testing if the question is even relevant to ask.

Thus, this study aims to answer these questions: \textit{can generative AI tools flag difficulties that may arise in translation of English survey questions? If so, what are the types of translation problems the AI identifies?}

\section{Methodology}\label{sec:methodology}

Users engage with ChatGPT web by writing and submitting prompts. To systematically assess the possibility of using ChatGPT to assist with survey translation, we utilize an experiment that allows us to systematically test prompt variations. We use 282 survey questions as experimental units in a zero-shot prompt experiment to understand the effect of generative AI model and target linguistic audience on the generative AI output. The number of questions was chosen based on three factors: the availability of questions that met our criteria, the need to minimize processing time, and the results of the power analysis presented in Appendix \ref{Appendix:poweranalysis}.

\subsection{Sources of Experimental Units}
We sourced 282 questions from surveys developed by Gallup, World Values Survey Association (WVS), and the Governance and Local Development Institute (GLD). From each organization we selected questions from surveys in the original English version that have been implemented in multi-cultural and multi-linguistic contexts requiring translation. From Gallup, we selected the publicly available Q12 survey containing 13 questions to measure employee engagement. Since its original release 20 years ago, the questions have been  been translated into at least 56 languages \autocite{gallup_overview}. We would have used questions from Gallup's other international surveys, but they were not publicly available. Additionally, we selected questions from the 7th wave of WVS. This survey has been translated and implemented in over 90 countries since the first wave in 1981 \autocite{wvs_methodology}. Finally, the 2019 version of the LGPI survey conducted by GLD was implemented in 3 countries requiring six translations \autocite{gld_overview}. 

Specific questions from Gallup, WVS, and LGPI were selected on grounds of their ability to be posed as ``stand alone" questions. Questions were excluded if they relied on answer set structure, survey logic, were not phrased as a question, or were redundant. Questions that were structured as fill in the blank (for example, ''I am...with my government.") were also excluded.

The researchers also wrote 10 ``questionable questions" (referred to as QQs henceforth) based on the survey development literature and 10 from the translation literature.\footnote{The first 10 were constructed for a separate experiment, but were included based on their generic nature. See Appendix \ref{Appendix:EU} for the full list of questions included in our sample.} Each question targets at least one specific problem. The QQs were purposefully culturally sensitive with concepts the literature identifies as difficult to translate like gender, medical terms, and age or syntactically difficult with additional clauses, double negatives, or formality issues. For example, "How many peers do you have at work?" (QQ13) contains the problem of inconsistent conceptualization identified specifically with the word ''peers" by Weeks et al (2007).

\subsection{Experimental Design}\label{subsec:design2}
We present a zero-shot prompt experiment to assess how different prompt structures impact the ability of generative AI models to identify components of survey questions that should be considered before translation. ``Zero-shot" indicates that we do not train or interact with the model, so that the model produces content based on its full training material. The treatment design, summarized in Table \ref{tab:experimentaldesign2}, is a 2$\times$3 factorial design resulting in 6 unique treatments. The two factors we randomize are: the version of GPT and the target linguistic audience of the translated survey stated in the prompt.

The treatments are realized in a prompt that is pasted into ChatGPT using the appropriate GPT model.\footnote{Full set of prompts available in Appendix \ref{Appendix:prompts}.} The general form of the prompt is: 

\begin{quote}
    ``Give me up to 5 aspects of this question that will be difficult to translate into \{\textbf{Language}\} for people living in \{\textbf{Location}\}. You are an expert on \{\textbf{Location}\}. \{\textbf{Question}\}"
\end{quote}
Note that the Target Linguistic Audience (TLA) factor provides both context ``\textit{{\textbf{Language}\} for people living in \{\textbf{Location}\}}}" and a persona ``\textit{You are an expert on \{\textbf{Location}\}.}" to the model. Providing the AI with context and a persona, is a well-documented method for obtaining higher quality output \autocite{white_prompt_2023}.

{
\renewcommand{\arraystretch}{0.8}
\begin{longtable}{|p{20em}|l|}
\caption{Summary of Randomized Factors and Corresponding Levels}
\label{tab:experimentaldesign2} \\

\hline \multicolumn{1}{|c|}{\textbf{Attribute}} & \multicolumn{1}{c|}{\textbf{Levels}} \\ \hline
\endfirsthead

\multicolumn{2}{c}%
{{\bfseries \tablename\ \thetable{} -- continued from previous page}} \\
\hline \multicolumn{1}{|c|}{\textbf{Attribute}} & \multicolumn{1}{c|}{\textbf{Levels}} \\ \hline 
\endhead

\hline \multicolumn{2}{|r|}{{Continued on next page}} \\ \hline
\endfoot

\hline \hline
\endlastfoot
    \textbf{Model }  &  GPT--3.5 \\ 
      &  GPT--4 \\ \hline
\textbf{Target Linguistic Audience (TLA) -} &    translated \\
 Language + Location &   Castillian Spanish in Spain\\
&   Mandarin Chinese in Mainland China\\ \hline
\end{longtable}
}

We randomize model between GPT--3.5 and GPT--4 because new versions of generative AI, including ChatGPT, are consistently being released. Little research exists to what extent these updates make a difference in output quality and usability for large data collections even though, it is widely assumed that the newer model should be better. Thus, it is important to keep tracking these updating versions to see any potential effects on output. 

We randomize the target linguistic audience to test how prompt construction may influence output. Prompts given to LLMs enforce rules, automate processes, and ensure specific qualities and quantities of generated output; it also helps the researcher reflect to the AI what the researcher wants specifically \autocite{white_prompt_2023}. Our control condition for TLA provides a baseline of what the AI can generate with fewer restrictions. Then we specify (1) Castilian Spanish in Spain or (2) Mandarin Chinese in Mainland China. Both of these target languages have large online presences\autocite{internetusage}. What is more, existing materials including academic articles, practitioner's experience, and online forums regarding translating from English into the two most widely spoken languages by native speakers should be extensive. We specify target location in addition to the language in order to minimize the amount of output that would mention the dialects of each language. 

We also provide two target groups to theoretically capture a broad range of issues affiliated with equivalency \autocite{behling2000translating}. Translating to Chinese faces more issues related to syntax and semantics compared to Spanish because English and Spanish belong to the same language family (Indo-European) with an alphabet whereas Chinese is Sino-Tibetan with characters. Many questions in our sample pertain to personal opinions and government which are especially susceptible to problems relating to context. It is fair to assume that translating from English to Mandarin for people in China (compared to Taiwan or Chinese speakers who live abroad) might be increasingly difficult to translate because some concepts related to, for example democracy or elections, may not be relevant or exist.  It is especially helpful if the AI is able to identify these non-existent concepts or potential sensitivity concerns. 

\subsection{Data Collection}\label{subsec:data}
The data collection for the experiment consisted of two steps: 1) obtaining output from the generative AI models, and 2) qualitatively coding the AI output. Practicality of the procedure is addressed in the results section. 

\subsubsection{Obtaining the GPT Output About the Survey Questions}
Each survey question in our sample serves as an experimental unit (EU) which received all treatments. We used an R script to create all experimental and formatting prompts.\footnote{Find all prompts in Appendix \ref{Appendix:prompts}.} Procedure \ref{procedure_summary1} describes the general process of obtaining the AI output. The final step was implemented using the R script available in the Harvard Dataverse repository (\cite{DVRepo}). 

\begin{algorithm}
\caption{Steps to Generate Experimental Output}\label{procedure_summary1}
\begin{algorithmic}
\State $n =$ number of survey questions in the sample
\State $t =$ number of unique treatments
\For{$treatment\gets 1, t$}
\For{$question\gets 1, n$}
\State Start a new GPT Chat
\State Enter the prompt as defined by the $treatment$
\State Enter the $question$
\State Enter the formatting prompt
\EndFor
\EndFor
\State Export all chats to a .json file
\State Convert .json to .xlsx file
\end{algorithmic}
\end{algorithm}

\subsubsection{Qualitative Coding of Generative AI Output}

The authors developed a codebook that primarily reflected translation issues from the literature but also left room for novel codes to emerge. This codebook was developed from the QQ output with two rounds of refinement and recoding. Table \ref{tab: codebook simple} is the simplified version of the codebook with code numbers that correspond with the results. \footnote{See Table \ref{tab:codescheme2} in Appendix \ref{appendix:qualcodingdetails} for the the complete coding guide including relevant literature and examples.} Each of these codes addresses one or many threats to equivalence according to Behling and Law (2000). We selected codes so that the primary researcher can decide how to proceed with the survey question and translator with more knowledge and awareness of potential problems.  

One of the emergent codes identified was when there was a problem in the source question that would be a problem regardless of and even with translation. For example, the AI was able to flag ''Do you think the job the police and courts are doing is excellent, good, fair or poor?" (QQ4) as a double barrelled question without asking for this type of feedback. We will not further address this code as it is not relevant to the research questions, but highlight it as an example of one of the emergent codes we include in the codebook.\footnote{Full results can be found in the Appendix \ref{Appendix:fullresults}} Another clarification as we coded output was a jointly accepted list of common source survey language for Code 1 found in Appendix \ref{Appendix:commonsourcelang}. 

\begin{table}
\centering
\caption{Qualitative Codebook (Simplified)} \label{tab: codebook simple}
\begin{tabular}[t]{|l|p{15em}|p{15em}|}
\hline
Code & Name & Description \\ \hline
Code1 & COMMON SOURCE SURVEY LANGUAGE & Words/Phrases commonly used in source language that are cumbersome or strange to translate \\ \hline
Code2 & TECHNICAL TERMINOLOGY &  Specialized terms \\ \hline
Code3 &  INCONSISTENT CONCEPTUALIZATION & When the concept is not universal; if it’s translatable but the answer will be unclear; may include internal inconsistency within the country/culture \\ \hline
Code4 & GENDERED LANGUAGE & When source question uses a gendered word and translators will not know whether to keep the gender; translated question has gender issues \\ \hline
Code5 & FORMALITY &  When the tone, politeness, or register may be relevant \\ \hline
Code6 &  SYNTAX &  If you can translate but it will not be smooth; awkwardness and clunky; problems affiliated with tenses \\ \hline
Code7 &  CULTURAL/REGIONAL SPECIFIC TERMS & Identify a word or phrase that exists in original language and a different, specific version exists in translation, may include regional variation \\ \hline
Code8 &  NONEXISTENT, IRRELEVANT, OR UNFAMILIAR CONCEPT & Topic or concepts that do not meaningfully exist or matter in translated language or culture \\ \hline
Code9 & SENSITIVE TOPIC &  Topic is emotional or taboo and may spark strong emotions \\ \hline
Code10 &  Excluded in Analysis & Problem affiliated with source language not translation \\ \hline
NOTA & NONE OF THE ABOVE & Non specific advice, nonsensical, seems like AI is “reaching” for 5 statements \\ \hline
\end{tabular}
\end{table}

\subsection{Analysis Plan}\label{subsec:analysisplan}
Our analysis proceeds as follows: the statistical analysis of pre--registered hypotheses and additional exploratory analyses.\footnote{All statistical analysis is performed using R (\cite{RSoftware}). Regression tables are produced using the \texttt{texreg} package (\cite{texreg}) and figures are produced using the \texttt{ggplot2} package (\cite{ggplot2}).}

\subsubsection{Analyzing Pre--Registered Hypotheses}
The following hypotheses are inherently exploratory and not based on previous research:\footnote{The hypotheses and analysis plan were submitted on OSF on November 23, 2023 \url{https://osf.io/4q26k}.} 

\begin{itemize}
    \item  H1 (Main Effect of Model): We expect that version 4 of GPT is more likely to flag all codes than version 3.5. 
    \item H2 (Effect of Target Linguistic Audience on "None of the Above"): The likelihood of flagging ''None of the Above" is lower when either target audience is specified compared to when it is unspecified. 
    \item H3 (Effect of Target Linguistic Audience on Specific Codes): Specifying the Target Linguistic Audience (Chinese or Spanish) will increase the likelihood of flagging the following codes compared to when the Target Linguistic Audience is not specified: Code 3: Inconsistent Conceptualization, Code 7: Cultural or Regional Terms, Code 8: Non-existent, irrelevant, or unfamiliar concept.
\end{itemize}

Given the factorial experimental design and that every experimental unit receives each treatment we use 2-level hierarchical regression models to test our pre--registered hypotheses. When the outcome is binary we fit the following model: 
\begin{equation}\label{eq:MLLM}
logit(y_{ij}) =  \beta_0 + \beta_M\mathbf{1}_{GPT-4} +  \beta_{T2}\mathbf{1}_{Spanish} + \beta_{T3}\mathbf{1}_{Chinese} + u_i + \varepsilon_{ij}
\end{equation}
where $y_{ij}$ takes on the value 0 or 1 indicating if a specific code was flagged for the output from question $i$ after receiving treatment $j$. The $\mathds{1}$s represent indicator variables for the noted factor level and the $\beta$s are the corresponding coefficients. The term $u_i$ is the random intercept associated with the question and $\varepsilon_{ij}$ represents the error. The inclusion of the random question intercept address the fact that each treatment was applied to every question.

We will also fit 2-level linear probability models for each binary outcome as shown in Equation \ref{eq:LPM}:
\begin{equation}\label{eq:LPM}
y_{ij} =  \beta_0 + \beta_M\mathbf{1}_{GPT-4} +  \beta_{L1}\mathbf{1}_{Spanish} + \beta_{L2}\mathbf{1}_{Chinese} + u_i + \varepsilon_{ij}. 
\end{equation}
When the results are robust for the multilevel logistic and multilevel linear probability models, we present the multilevel linear probability models due to their ease of interpretation.\footnote{The full results will be available in Appendix \ref{Appendix:robustness}.}

\subsubsection{Exploratory Analyses}

Additionally, we present a series of exploratory analyses, some of which were suggested as such in the pre--analysis plan and others that emerged after qualitative analysis. First, we analyze the effect of question source on total number of codes and likelihood of flagging individual codes. Next, we test interaction effects between model and target linguistic audience; the initial power analysis suggested we would be under powered to run this analysis, but effect sizes were much larger than expected. Both analyses will utilize modified version of the 2-level hierarchical models presented in Equations \ref{eq:MLLM} and \ref{eq:LPM}. 

Finally, as referenced in the pre-analysis plan, we explore if there is any structure or information contained in the placement of the flagged codes, i.e, are certain codes more likely to appear earlier or later in the 5 statements? We run a one-way ANOVA test to determine if all codes have the same average statement placement. Should we find a statistically significant result, we run a Tukey HSD test to determine which codes have similar and different average placement. 

\section{Results}
\subsection{Overview of Data}\label{sec:overview}

We qualitatively coded a total of 8460 statements.\footnote{282 questions x 6 treatments x 5 statements per treatment = 8460 statements} Each statement had the possibility of flagging any of the 9 primary codes shown in Table \ref{tab:codescheme2}. If the AI flagged no codes, the statement is considered NOTA (none of the above). The vast majority of the time (77\%) each statement is flagged with only one code. About 10\% of the time 2 codes are flagged, 13\% of the time no codes were flagged and less than 0.5\% of the time three codes are flagged. The AI was most likely to flag Code 3 -- Inconsistent Conceptualization (over 35\%). Non-existent or unfamiliar concepts (Code 8 - 13\%) and none of the above (NOTA - nearly 13\%) were the second and third most frequent codes respectively. All other codes constituted less than 7.5\% of all coded statements.\footnote{See Figure \ref{fig:DistCode_Statement} in Appendix \ref{Appendix:fullresults} for the distribution of codes at the statement level.}

\begin{figure}[htbp]
    \centering
    \caption{Distribution of the Number of Codes at the Treatment-Question Level}
    \label{fig:DistNumCode_TrtQ}
    \includegraphics[width = 0.8\linewidth]{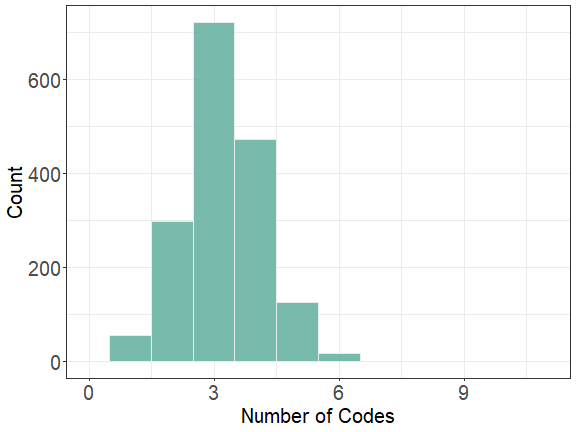}
    
\end{figure}

We also consider the data at the treatment-question level by creating indicator variables representing the presence of each code in any of the five statements, resulting in 1692 unique observations. Figure \ref{fig:DistNumCode_TrtQ} shows that the distribution of flagged codes per treatment-question pair is approximately normal ranging from 1 to 6. We see that inconsistent conceptualization (Code 3) is nearly 25\% of all flagged codes; this decreased compared to the statement level. Notably, none of the above is coded nearly 13\% of the time, as the second most likely to be identified code at the question-treatment level. Non-existent or unfamiliar concepts (Code 8) and sensitive material (Code 9) are 13\% of the flagged codes respectively. All other codes are flagged less than 7.5\% of the time.

\subsection{Pre--Registered Hypotheses}\label{sec:prereghypo}

All observations are analyzed at the treatment-question pair level. Presented results are multilevel linear probability model (MLLPM) results unless explicitly stated to be multilevel logistic model (MLLM) results.\footnote{Full results, MLLPM and MLLM, are available in Appendix \ref{Appendix:fullresults}}

\subsubsection{Model Effect}
We find that the relationship between model and presence of codes is more diverse than expected. Table \ref{tab:NumCodesTrtQ} in Appendix \ref{Appendix:fullresults} shows that the average number of total codes flagged by GPT--4 is 0.10 less ($p<0.05$) than GPT--3.5. 

\begin{figure}
    \centering
    \caption{95\% Confidence Intervals of the Model Effect of the Likelihood of Flagging Each Code}
    \label{fig:ModelEffectPlot}
    \includegraphics[width=0.8\linewidth]{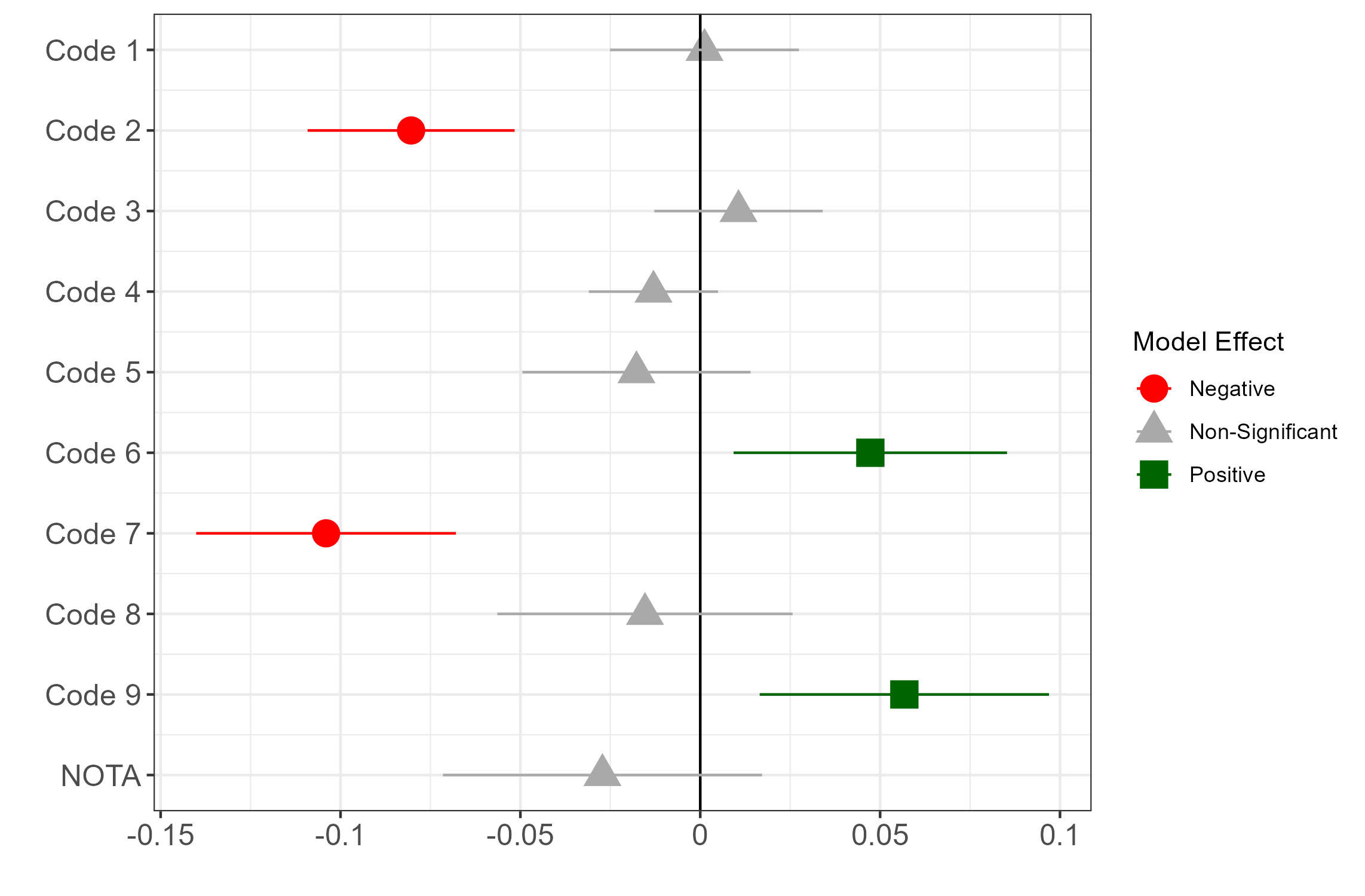}
\end{figure}

Figure \ref{fig:ModelEffectPlot} summarizes the model effect of using GPT--4 compared to GPT--3.5 on the likelihood of flagging the 10 codes (including NOTA) and Table \ref{tab:RegCode1to10} provides the full results (excluding NOTA). When using GPT--4.0 compared to GPT--3.5, the likelihood of flagging a syntax issue (Code 6) increases by 0.05 ($p<$0.05) on average and the likelihood of flagging sensitivity issues (Code 9) increases 0.06 ($p<$0.01) on average. Whereas, for Code 2 and Code 7 we find the opposite effect. When using GPT--4.0 compared to GPT--3.5,  the likelihood of flagging technical terms (Code 2) decreases  by 0.08 ($p<$0.001) on average and for cultural or regionally specific terms (Code 7) the likelihood decreases 0.10 ($p<$0.001) on average. We find no statistically significant evidence of a model effect on the other codes. 

\subsubsection{Effect of Target Linguistic Audience}

Table \ref{tab:RegNOTA} in Appendix \ref{Appendix:fullresults} shows that when specifying the target linguistic audience as Spain compared to no linguistic audience, the likelihood of flagging NOTA increases by 0.32 ($p<$0.001) on average. Likewise, specifying China increases the likelihood by 0.16 ($p<$0.001) on average. 

\begin{table}[htbp]
\begin{center}
\caption{Multilevel Regression Results for Codes 1 to 10}
\label{tab:RegCode1to10}
\begin{tabular}{l c c c c c}
\hline
 & Code 1 & Code 2 & Code 3 & Code 4 & Code 5 \\
\hline
(Intercept)               & $0.30^{***}$  & $0.21^{***}$  & $0.96^{***}$  & $0.07^{***}$ & $0.07^{***}$ \\
                          & $(0.02)$      & $(0.02)$      & $(0.01)$      & $(0.01)$     & $(0.02)$     \\
ModelM2                   & $0.00$        & $-0.08^{***}$ & $0.01$        & $-0.01$      & $-0.02$      \\
                          & $(0.01)$      & $(0.01)$      & $(0.01)$      & $(0.01)$     & $(0.02)$     \\
AudienceT2                & $-0.05^{**}$  & $0.04^{*}$    & $-0.08^{***}$ & $0.04^{**}$  & $0.16^{***}$ \\
                          & $(0.02)$      & $(0.02)$      & $(0.01)$      & $(0.01)$     & $(0.02)$     \\
AudienceT3                & $-0.09^{***}$ & $0.01$        & $-0.06^{***}$ & $-0.01$      & $0.14^{***}$ \\
                          & $(0.02)$      & $(0.02)$      & $(0.01)$      & $(0.01)$     & $(0.02)$     \\
\hline
AIC                       & $1119.62$     & $1216.68$     & $344.91$      & $-310.38$    & $1323.42$    \\
BIC                       & $1152.22$     & $1249.29$     & $377.52$      & $-277.78$    & $1356.02$    \\
Log Likelihood            & $-553.81$     & $-602.34$     & $-166.46$     & $161.19$     & $-655.71$    \\
Num. obs.                 & $1692$        & $1692$        & $1692$        & $1692$       & $1692$       \\
Num. groups: Question     & $282$         & $282$         & $282$         & $282$        & $282$        \\
Var: Question (Intercept) & $0.11$        & $0.06$        & $0.02$        & $0.03$       & $0.02$       \\
Var: Residual             & $0.08$        & $0.09$        & $0.06$        & $0.04$       & $0.11$       \\
\hline
\\ \\
\hline
 & Code 6 & Code 7 & Code 8 & Code 9 & Code 10 \\
\hline
(Intercept)               & $0.22^{***}$ & $0.23^{***}$  & $0.68^{***}$  & $0.35^{***}$ & $0.16^{***}$ \\
                          & $(0.02)$     & $(0.02)$      & $(0.02)$      & $(0.02)$     & $(0.02)$     \\
ModelM2                   & $0.05^{*}$   & $-0.10^{***}$ & $-0.02$       & $0.06^{**}$  & $0.01$       \\
                          & $(0.02)$     & $(0.02)$      & $(0.02)$      & $(0.02)$     & $(0.02)$     \\
AudienceT2                & $0.05$       & $0.19^{***}$  & $-0.34^{***}$ & $-0.04$      & $-0.04$      \\
                          & $(0.02)$     & $(0.02)$      & $(0.03)$      & $(0.03)$     & $(0.02)$     \\
AudienceT3                & $-0.01$      & $0.06^{**}$   & $-0.20^{***}$ & $0.33^{***}$ & $-0.00$      \\
                          & $(0.02)$     & $(0.02)$      & $(0.03)$      & $(0.03)$     & $(0.02)$     \\
\hline
AIC                       & $1930.02$    & $1836.83$     & $2237.75$     & $2167.26$    & $1331.45$    \\
BIC                       & $1962.63$    & $1869.43$     & $2270.35$     & $2199.86$    & $1364.05$    \\
Log Likelihood            & $-959.01$    & $-912.41$     & $-1112.87$    & $-1077.63$   & $-659.73$    \\
Num. obs.                 & $1692$       & $1692$        & $1692$        & $1692$       & $1692$       \\
Num. groups: Question     & $282$        & $282$         & $282$         & $282$        & $282$        \\
Var: Question (Intercept) & $0.03$       & $0.04$        & $0.05$        & $0.04$       & $0.01$       \\
Var: Residual             & $0.16$       & $0.14$        & $0.19$        & $0.18$       & $0.12$       \\
\hline
\multicolumn{6}{l}{\scriptsize{$^{***}p<0.001$; $^{**}p<0.01$; $^{*}p<0.05$}}
\end{tabular}

\end{center}
\end{table}

Table \ref{tab:RegCode1to10} shows the full main effects results for the likelihood of flagging each primary code. For many codes we see a similar target linguistic audience effect regardless of whether Spain or China is specified, though we find relationships in both the positive and negative directions. Regarding positive effects, we find that the average likelihood of flagging formality concerns (Code 5) increases by 0.16 ($p<$0.001) and 0.14 ($p<$0.001) for Spain and China respectively. Similarly, the average likelihood of flagging cultural or regional terms (Code 7) increases by 0.19 ($p<$0.001) and 0.06 ($p<$0.01) for Spain and China respectively. 

Regarding consistent negative effects, the average likelihood of flagging common survey language (Code 1) decreases by 0.05 ($p<$0.01) and 0.09 ($p<$0.001), inconsistent conceptualization (Code 3) decreases by 0.08 ($p<$0.001) and 0.06 ($p<$0.001) , and non-existent or irrelevant concept (Code 8) decreases by 0.34 ($p<$0.001) and 0.20 ($p<$0.001) each for Spain and China respectively. 

For codes 2, 4, and 9, we find a significant effect for one audience but not the other. We find that average likelihood of flagging technical terminology (Code 2) and for flagging gendered language (Code 4) both increase by 0.04 ($p<$0.05 for Code 2, $p<$0.01 for Code 4) when Spain was specified, but we find no statistically significant effects when China is specified. Whereas, we find that the average likelihood  of flagging the presence of sensitive topics (Code 9) increases by 0.33 ($p<$0.001) when China is specified, and no evidence of an effect when Spain is specified. Syntax issues (Code 6) is the only code for which we find no target linguistic audience effect.  

\subsection{Exploratory Analyses}\label{sec:exploratory}

We also conducted three exploratory analyses to provide more insights into the viability of incorporating generative AI into survey translation practices. Specifically we run analyses to test for question source effects, interaction effects between the experimental attributes Model and Target Linguistic Audience, and an analysis on the order that codes appeared in the 5 statements produced. 

\subsubsection{Question Source Effect}

We find that the question source did have a significant effect on the average number of codes produced. Table \ref{reg:Source} shows that the average number of codes per question--treatment for questions from the LGPI was not statistically different from the questionable questions. But the average number of codes for the World Values Survey's questions decreases by  0.24 ($p<$0.05) compared to the questionable questions, and average number of codes for Gallup's questions decrease by 0.61 compared to the questionable questions when controlling for model and target linguistic audience. 

Given that the three question sources included in this work are merely a subset of all possible question sources, it would be reasonable to model the data in a 3-level structure of measurement within question within source.\footnote{As a robustness check we provide versions of all analyses using a 3-level structure in Appendix \ref{Appendix:robustness}. We note that all findings are robust to a 2-level or 3-level specification.} The significance of the question source is also supported by the non-zero (albeit small) variance of the random question source intercept in the three-level hierarchical model presented in Table \ref{reg:3MLMSource}. 

\subsubsection{Interaction Effects}
We were able to check for interaction effects between Model and Target Linguistic Audience because we found main effect sizes larger than estimated in the power analysis. We find evidence of a statistically significant interaction on the total number of codes and on the likelihood of flagging 4 of the 9 primary codes. Note that for the interaction models regarding the likelihood of flagging codes we refer to the multilevel logistic regression results due to a lack of consistency with the multilevel linear probability models. 

Table \ref{tab:RegNumCodesInt} accompanied by Figure \ref{fig:NumCodesIntplot} in Appendix \ref{appendix:interactions} shows that the effect of Model on the total number of codes is not consistent across Target Linguistic Audience. When no target is specified (T1), we find no evidence of a Model effect, but when Spain (T2, $p<0.01$) or China (T3, $p<0.05$) is specified, the effect of using GPT--4 is less than the effect of using GPT--3.5 with the effect more pronounced when Spain is specified. 

\begin{figure}[!t]
\caption{Predicted Probabilities from Hierarchical Logistical Regression Results - Statistically Significant Model by Target Audience Interaction Effects on the Likelihood of Flagging of Codes 5, 7, 9, and NOTA. }
  \label{fig:StatSigInt}
  \centering
  \begin{subfigure}[b]{0.45\textwidth}
    \includegraphics[width=\textwidth]{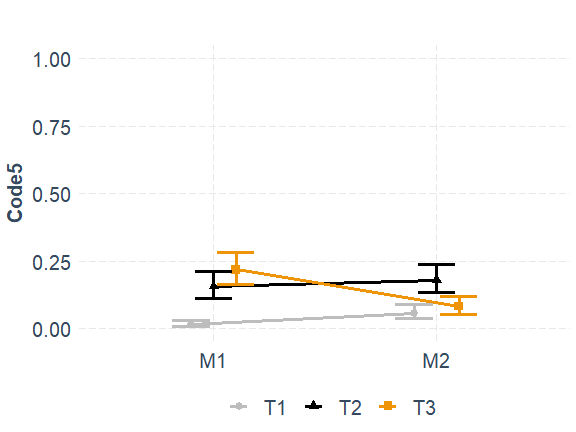}
      \captionsetup{width=.8\linewidth}
    \caption{Interaction Plot for the Interaction of Model and Target Audience on the Presence of Code 5}
    \label{fig:Code5intplot}
  \end{subfigure}
  \begin{subfigure}[b]{0.45\textwidth}
    \includegraphics[width=\textwidth]{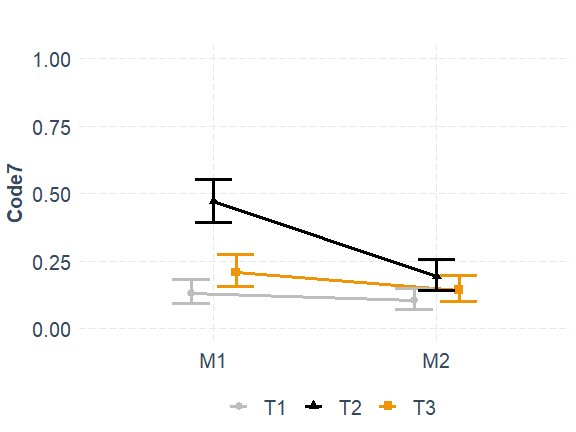}
      \captionsetup{width=.8\linewidth}
    \caption{Interaction Plot for the Interaction of Model and Target Audience on the Presence of Code 7}
    \label{fig:Code7intplot}
  \end{subfigure}
  \begin{subfigure}[b]{0.45\textwidth}
    \includegraphics[width=\textwidth]{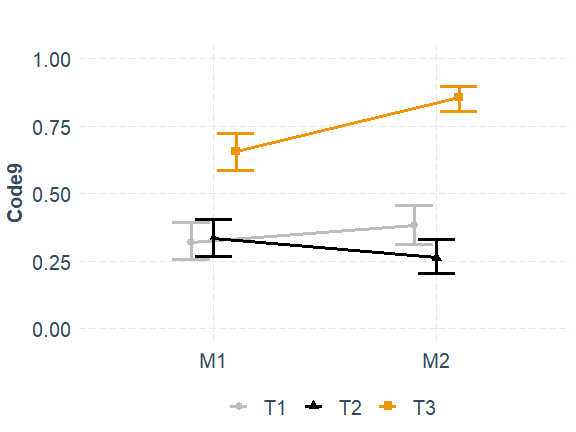}
      \captionsetup{width=.8\linewidth}
    \caption{Interaction Plot for the Interaction of Model and Target Audience on the Presence of Code 9}
    \label{fig:Code9intplot}
  \end{subfigure}
  \begin{subfigure}[b]{0.45\textwidth}
    \includegraphics[width=\textwidth]{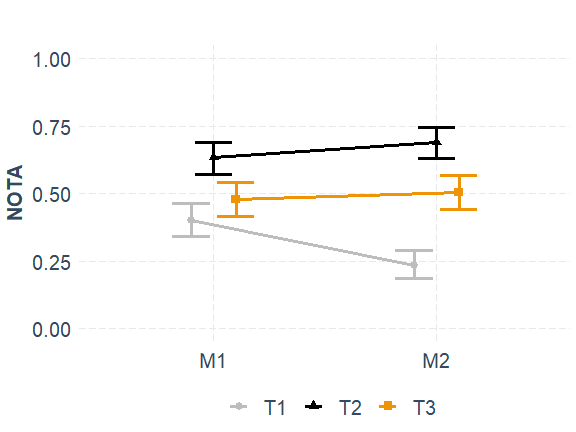}
      \captionsetup{width=.8\linewidth}
    \caption{Interaction Plot for the Interaction of Model and Target Audience on the Presence of Code NOTA}
    \label{fig:CodeNOTAintplot}
  \end{subfigure}
  
\end{figure}

Figure \ref{fig:StatSigInt} presents interaction plots for the four codes with a significant Model by Target Linguistic Audience interaction.\footnote{We also found evidence of a statistically significant interaction for Code 4; however, Table \ref{tab:RegCode1to5log.int} and Figure \ref{fig:Likely4Intplot} show that the evidence is weak $p=0.0318$ and the effect size is very small. Thus we exclude this result from our presentation.} We find that the effect of model is not consistent across target linguistic audience for formality issues (Code 5), cultural or regionally specific terms (Code 7), sensitive topics (Code 9), and none of the above (NOTA).\footnote{Full results in Tables \ref{tab:RegCode1to5log.int}, \ref{tab:RegCode6to10log.int}, and \ref{tab:RegNOTAlog.int} in Appendix \ref{Appendix:robustness}.} Furthermore, the nature of each interaction is different for each code. 

Subfigure \ref{fig:Code5intplot} shows that the Model and Target Linguistic Audience effects are much lower for formality issues (Code 5) than the other three codes with the predicted probabilities ranging from close to 0 to 25\%. When no TLA is specified we see the likelihood of flagging Code 5 is greater for GPT--4. Comparatively, when Spain ($p<0.05$) is specified, the likelihood is much higher than when no TLA is specified, but we find no evidence of a difference between the effect of the two GPT versions.  Alternatively, when China ($p<0.001$) is specified we find that the likelihood of flagging formality issues is much lower when GPT--4 is used compared to GPT--3.5. 
Subfigure \ref{fig:Code7intplot} shows that the likelihood of flagging cultural/regional specific terms (Code 7) is greatest when Spain is specified, with the likelihood much lower when GPT--4 is specified ($p<0.001$). There appears to be no model effect when no TLA is specified or when China is specified. 

Subfigure \ref{fig:Code9intplot} shows that we find different model effects for all three TLAs. We see that the likelihood of flagging sensitivity issues (Code 9) is greatest when China ($p<0.01$) is specified as the TLA, with the effect magnified when GPT--4 is used. The likelihood is much lower when Spain or no TLA is specified. However, when Spain is specified, we find weak evidence ($p<0.05$) that the likelihood is lower when GPT--4 is specified. 
Subfigure \ref{fig:CodeNOTAintplot} shows that when no TLA is specified, GPT--4 is less likely to flag NOTA than GPT--3.5 (M2 effect = -0.79, $p<$0.001). In contrast, when Spain ($p<0.001$) or China ($p<0.001$) is specified as the TLA, we find no evidence of a model effect, but the over likelihood of flagging a NOTA is higher for Spain than China. 

\subsubsection{Statement Order Analysis}

\begin{figure}[htbp]
    \centering
    \caption{Average Statement Placement by Code, Sorted from Least to Greatest and Labelled with Groupings from Tukey HSD Test}
    \label{fig:StatementOrder}
    \includegraphics[width = 0.7\linewidth]{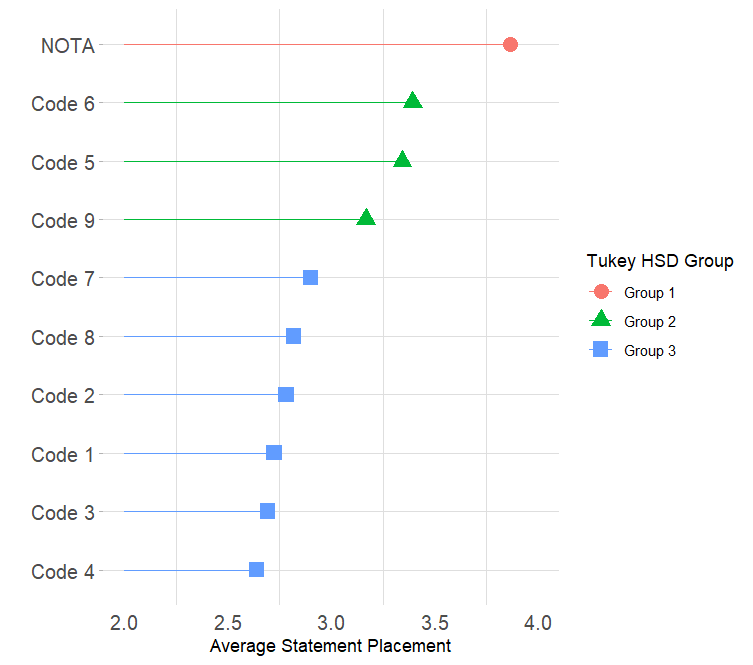}
    
\end{figure}

To better understand the previously presented results, we checked if there were any meaningful patterns in the order by which the AI flags certain types of problems. Figure \ref{fig:StatementOrder} shows the average statement placement (1--5) of each code. The one--way ANOVA test ($p<0.001$, full results in Table \ref{tab:ANOVA}) shows that the average placement is not the same for all codes. 

To uncover which codes have similar and different average placement, we perform a Tukey HSD test. The findings of the test are summarized in Figure  \ref{fig:StatementOrder}.\footnote{Full Tukey HSD results available in Table \ref{tab:TukeyHSD} in Appendix \ref{Appendix:fullresults}} Specifically, we see that NOTA has a different (higher) mean than all other codes (mean = 3.8). Meaning that on average, statements flagged as NOTA appear later in the statement order. The remaining codes are clustered in two groups. Formality (Code 5), Syntax (Code 6), and Sensitivity (Code 9) are one group with averages just below the average placement of NOTA (means ranging from 3.1 to 3.4) and all the remaining codes are in the second group (means ranging from 2.6 to 2.9).

\subsection{Procedural and Logistical Analysis}\label{sec:proclogi}
This section highlights the added value of ChatGPT as a tool, compared to preparing surveys for translation without AI--especially under limited time and funding constraints. First, we show ChatGPT's potential capability compared to a human; second, we highlight the logistics from our specific experiment. This is relevant for those who plan to use ChatGPT to supplement other translation tools and also for a researcher planning to run a similar computational experiment generating this much output from ChatGPT.

\subsubsection{Added Value of the AI}

\begin{figure}[htbp]
    \centering
    \caption{Number of Codes Flagged by Each Treatment}
    \label{fig:CodeCounts}
    \includegraphics[width=0.85\linewidth]{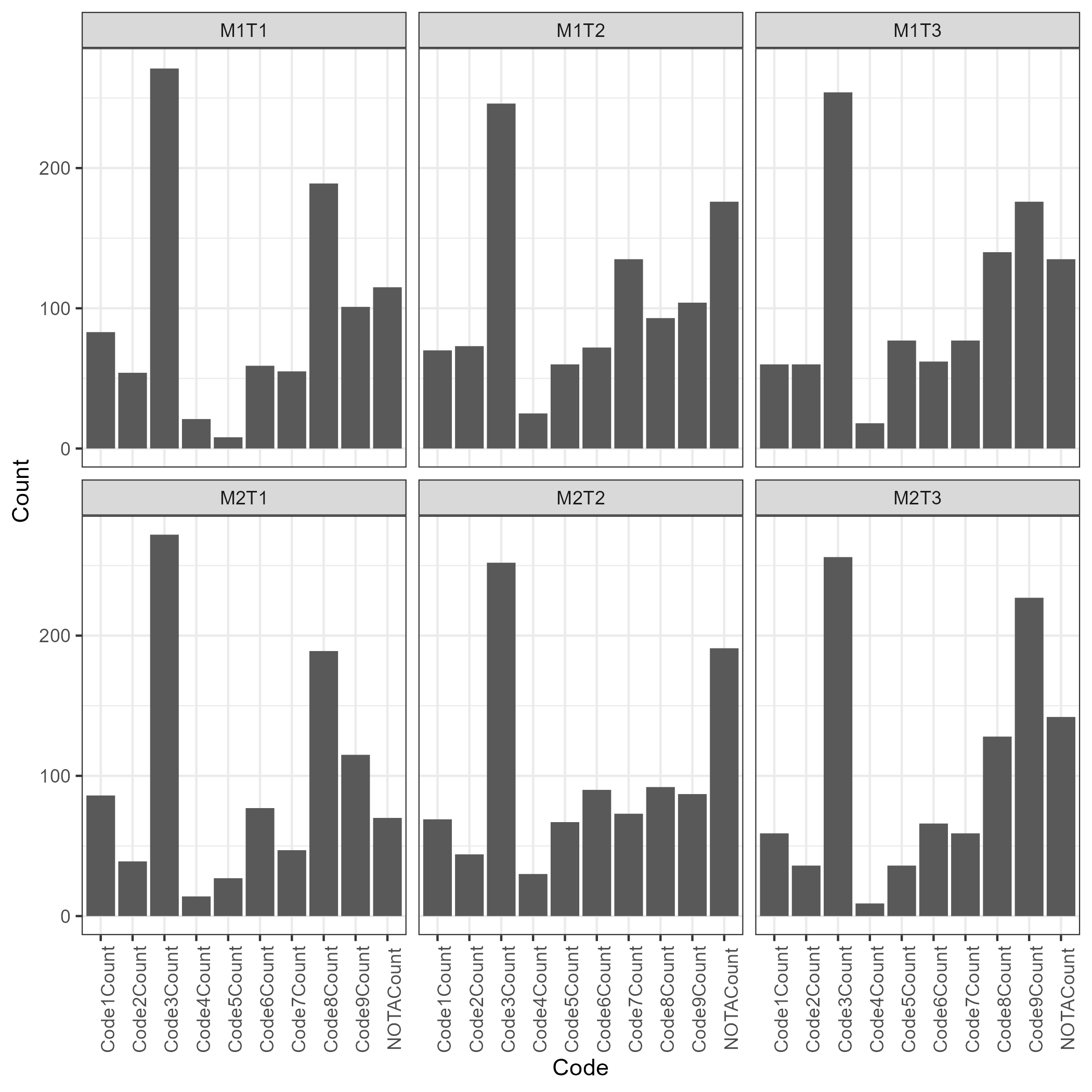}
\end{figure}

Although we cannot fully address concerns about output accuracy within the scope of this paper, we have reason to believe a follow-up study would be appropriate. Figure \ref{fig:CodeCounts} shows that all treatments produced statements that reflected the nine primary codes and the none of the above. 

This indicates that our prompts do not constrain the AI to produce only a subset of relevant codes. Furthermore, without training, the generative AI can produce statements with content relevant in the translation literature. 

While the authors do not possess the cultural or linguistic expertise to assess the accuracy of the AI output, a limited indication of quality is whether the AI can identify the problems in the questionable questions numbered 11 to 20 that were based on problems from the surveys in translation literature. Table \ref{tab:RecoverQQ} summarizes the number of treatments out of six that recovered the known codes. We see that all six treatments always returned Codes 2, 8, and 9. They also almost always marked Code 3. We see the lowest levels of recovery when there were two known codes. This is a very limited and narrow test of accuracy since each code, except Code 3, is evaluated using a single question. 

\begin{table}[htbp]
\centering
\caption{Distribution of Treatments that Recovered All Known Codes}
\label{tab:RecoverQQ}
\begin{tabular}{|l|l|p{7em}|p{7em}|}
\hline
Question & Known Code  & Not Accurately Recovered & Accurately Recovered\\
\hline
QQ11 & Code 1, Code 6 & 5 & 1\\
\hline
QQ12 & Code 3 & 5 & 1\\
\hline
QQ13 & Code 3 & 0 & 6\\
\hline
QQ14 & Code 3 & 2 & 4\\
\hline
QQ15 & Code 9 & 0 & 6\\
\hline
QQ16 & Code 2 & 0 & 6\\
\hline
QQ17 & Code 8 & 0 & 6\\
\hline
QQ18 & Code 4, Code 7 & 6 & 0\\
\hline
QQ19 & Code 5 & 1 & 5\\
\hline
QQ20 & Code 3 & 0 & 6\\
\hline
\end{tabular}
\end{table}

\subsubsection{Logistics}
At the time of data collection, we relied primarily on two premium accounts which cost 20 USD/month. Premium accounts permitted 50 prompts every 3 hours for model 4 and unlimited messages for model 3.5. We used a premium account to collect almost all data for model 3.5. Collecting 2 complete experimental units under all conditions (so 12 experimental prompts and 12 formatting prompts) took 10 minutes with GPT--4 and 5 minutes with GPT--3.5. The time between requesting data output and receiving the email with the .json file was 3-5 minutes. Running the R script and saving data in the proper files took an additional 3 minutes. Notably, the timing fluctuated depending on the demand and use worldwide so that it would be slower in peak business times for North America and Europe--especially when those times overlapped. 

The free version ChatGPT did not have unlimited access to GPT--3.5 as expected. Not publicly displayed by the company but corroborated in forums, the free version only produced output for about 50 entries (sometimes more or fewer depending on demand), then would require data collectors to wait one hour to continue. Additionally, the wait time from requesting the data output to receiving the email was typically closer to 10 minutes and sometimes hours as opposed to 5 minutes or less with the premium account. There appeared no meaningful differences in quality between the free and premium accounts for model 3.5, though this is beyond the scope of our paper. Given these restraints and no apparent quality, we relied on the paid version for model 3.5 for the majority of our data collection.

\section{Discussion}

Our experiment and subsequent analyses provide many insights into the potential of using AI to assist with survey translations as well as many avenues for future work. 

\subsection{Sensibility and Structure of AI Output}
The choice to ask for 5 aspects in the prompt was an arbitrary decision to set a limit on processing time and the volume of the output. However, it appears that five aspects is an appropriate number to request from the AI. In Subsection \ref{sec:overview} we saw that each statement generally flagged one code and the distribution of the number of codes across all 5 statements was approximately normally distributed with a mean of 3.2 and a standard deviation of 0.92. Thus asking for fewer than 5 aspects could result in less information than available and asking for more than 5 aspects could lead to unnecessary additional processing time and NOTA (None of the Above) coded statements. 

The Statement Order Analysis showed that the order that the content was presented was not completely random. We found that the AI is most likely to provide statements about common source survey language (Code 1), technical terminology (Code 2), inconsistent conceptualization (Code 3), gendered language (Code 4),regionally specific terms (Code 7), and irrelevant/non-existent concepts (Code 8) first, then formality (Code 5), syntax (Code 6), and sensitive content (9), and lastly NOTA. It appears that the AI produces NOTA statements when it ``runs out" of content as opposed to producing empty statements at random.

\subsection{Assessing Model Effects}
The version of ChatGPT used does have an impact on the output provided; however, the effect is not uniform and is more nuanced that expected. While we expected GPT--4 to be more likely to flag all codes (H1), we find that GPT--4 provides marginally fewer codes per question than GPT--3.5 on average. Additionally, GPT--4 is more likely to flag syntax (Code 6) and sensitivity issues (Code 9) and less likely to flag technical terms (Code 2) and regionally specific terms (Code 7) than GPT--3.5. 

The variation in model effect size and direction highlights the criticality of monitoring and evaluating model effects on specific tasks as we integrate generative AI into our routines. This is particularly important as generative AI research evolves and new models are released. For example, while only GPT--3.5 and GPT--4 were available during data collection, now GPT--4o is available for the free and premium account. 

\subsection{Importance of Context and Persona in Prompts}
Prompt engineering research establishes that the context and persona, denoted \textit{target linguistic audience (TLA)} in our experiment, given to an AI has a large impact on the output; our work is no exception. However, we find evidence that \textit{how} context and persona matters is highly variable. 

We anticipated that the likelihood of NOTA codes would be lower when a target audience is specified, but we find strong evidence to the contrary, and an even more nuanced finding when we consider the interaction effect between Model and TLA. Looking at Subplot D of Figure \ref{fig:StatSigInt} we see no evidence of a model effect when Spain or China is specified, though the overall likelihood of flagging a NOTA code is slightly higher for Spain than China. When no specific context is provided, the likelihood of flagging NOTA is lower for GPT--4 than GPT--3.5. That is, when no context is given, the newer model is less likely to produce empty statements. 

We also expected that the likelihood of flagging inconsistent conceptualization (Code 3), cultural or regional terms (Code 7), and non-existent/irrelevant/unfamiliar concepts (Code 8) would be greater when a TLA is specified (H3). Again we find that the reality is more diverse than our expectations. From Table 2, we find when a TLA is specified the likelihood of flagging Code 7 increases, but the likelihood of flagging Code 3 and Code 8 decrease. When a TLA is not specified, it is reasonable that the AI can essentially always state that inconsistent conceptualization or non-existent irrelevant concepts might be an issue. In some ways, this may be seen as a trivial or softball response. The consistent finding for Code 7 is made more complex by the statistically significant interaction between Model and TLA on the likelihood of flagging Code 7. In this case, the interaction is driven by the case when Spain is specified and the likelihood of flagging Code 7 is lower for GPT--4 than GPT--3.5. Interesting we find little to no evidence of a model effect when China or no TLA is specified.

\subsection{Practicality \& Logistics}

From a financial point of view, the price of 20 USD/month is relatively low for access to newer GPT models. Furthermore, the free version appeared to be more limited in the volume it could produce even if the 3.5 model was not obviously a better model in all cases. Thus, we recommend the paid version, but one should consider their research task and time and personnel limitations.

Running similar computational experiments to the one presented here takes time but few specialized skills. A power analysis to estimate the required number of experimental units can be conducted using the effect sizes from our experiment\footnote{The code we used to perform our power analysis are available in the Harvard Dataverse \autocite{DVRepo}.}. Then, one needs to design prompts and copy and paste those prompts and the desired experimental units (survey questions) into the ChatGPT model. The largest logistical barrier is exporting and formatting the data, though this process can be automated and is not strictly necessary.\footnote{We provide an R script to format the data into an Excel file in the Harvard Dataverse (\cite{DVRepo}). Note that changes in the ChatGPT code may require modifications to the script, but it should serve a a good starting point.} The largest investment is the time to qualitatively code the GPT output, though this could possibly be streamlined using large language model (LLM) classifiers. 

From an application point of view, once robust and reliable prompts have been established, researchers will be able to simply feed their prompts and survey questions to the AI model and get the output. This is a task that can be easily delegated or even potentially automated. It will then be up to the researchers to determine how to proceed with the output  from ChatGPT.

\subsection{Outlining Future Research}

Our results suggest that there is great potential for AI to support survey translation best practices. However, before AI can be truly integrated, there are many questions to answer. Perhaps the greatest of these is the accuracy of the output generated by the AI. Given that we want the AI to be able to identify codes that a researcher would miss or be able to compensate for a lack of expertise in a variety of situations, an assessment of accuracy is critical before this procedure can be implemented. 

Even if it is determined that the AI output is sufficiently accurate, there are a host of follow up questions to interrogate. First, additional research on the ability to treat an entire survey questionnaire/battery at one time as opposed to our method of treating each question individually, could greatly reduce the data collection time. Furthermore, providing the AI with the full context of the survey could have a meaningful impact on the type and quality of the output. Second, a follow up study should evaluate the ability of using generative AI to assist with translating into languages that are not as widely spoken with online presences. Third, one could test few-shot prompting versus the zero shot prompting in our experiment in order to better fine-tune the types of output ChatGPT can provide.

Additionally, our work does not speak to how the AI output should be incorporated into translation practices. One could imagine using the GPT output to develop an annotated questionnaire that is given to the translators at the beginning of the Translation phase of the TRAPD method. This process could help the researcher be less reliant on the translators' knowledge and expertise by improving the quality of translations brought to the Review phase, to which most teams are able to devote limited time.  However, one might be concerned that providing the GPT output to the translators during the Translation phase might overly sway or guide the translators' thinking, missing items they may have otherwise flagged. If this is the case, one might suggest using the GPT output to the Review stage as a final check to issues raised by the translators independently of the GPT input. 

\section{Conclusion}

All institutions--academic, corporate, or otherwise--know how time, financial, and personnel constraints impact their ability to achieve their goals at the quality they desire. Therefore, everyone is searching for methods to overcome these them. The familiar chat format of ChatGPT makes generative AI a powerful tool accessible to a wide range of users. It also makes it tempting to open a new chat any time we have a problem to address, often without empirical evidence to justify the temptation. 

This study begins to fill a gap regarding the role generative AI might have in survey translation, investigating not how AI may create new best practices but how it can support and enhance existing ones. The current state-of-the-art method, TRAPD\footnote{Recall TRAPD stands for Translation Review Adjudication Pre-test Documentation}, has been shown to produce high-quality translations, but it requires extensive resources in the form of time, money, and expertise. Our zero-shot prompt experiment and subsequent analyses have shown that without any training ChatGPT models have the ability to provide feedback on a wide range of issues from the survey translation literature in a cost effective and time efficient manner. We also show that the construction of prompts and choice of AI model have significant impacts on the output generated and should be considered integral parts of future research in this area.

Our experiment suggests that there is great potential for generative AI to minimize survey error at the translation stage of 3M survey development. While this potential addition to best practices appears exceptionally useful in resource constrained scenarios, generative AI can help minimize cognitive burden and allow researchers and experts to spend more time solving problems than identifying them, even for those with near limitless resources. Thus, future research should continue to investigate and delineate the role that generative AI has to play within survey translation best practices.


\newpage
\printbibliography

\newpage
\setcounter{page}{1}
\appendix 
\appendixpage

\begin{appendices}

\section{Survey Questions}\label{Appendix:EU}
The following are the selected survey questions used as the experimental units. The label corresponds to the source of the question: QQ for ``Questionable Question," WVS for Wave 7 of the World Values Survey, G for Gallup's Employee Engagement Survey, and LGPI for Local Governance Performance Indicators from the Governance and Local Development Institute. 
\begin{longtable}{p{5em}|p{25em}}
\caption{Experimental Units} \label{appendixtable:EU}\\
\renewcommand{\arraystretch}{1.3}
\textbf{Label} & \textbf{Complete Question}                                                                                                                                                                                                                                                                                                                                                                                                                                                                             \\
QQ1            & Do   you think that adults should be able to use PCP without any legal penalty?                                                                                                                                                                                                                                                                                                                                                                                                                        \\
QQ2            & Before you got married, how long did you   live in Maryland after you graduated from college?                                                                                                                                                                                                                                                                                                                                                                                                          \\
QQ3            & How   long have you lived in College Park?                                                                                                                                                                                                                                                                                                                                                                                                                                                             \\
QQ4            & Do you think the job the police and   courts are doing is excellent, good, fair or poor?                                                                                                                                                                                                                                                                                                                                                                                                               \\
QQ5            & Please   tell me whether you agree or disagree with the following statement about   teachers in public schools: teachers should not be required to supervise   students in the halls, the lunchroom, and school parking lots                                                                                                                                                                                                                                                                           \\
QQ6            & Do you think women and children should   be given the first available flu shots?                                                                                                                                                                                                                                                                                                                                                                                                                       \\
QQ7            & Do   you support president Reagan’s policy on Nicaragua?                                                                                                                                                                                                                                                                                                                                                                                                                                               \\
QQ8            & Compared to a year ago, do you feel the   prices of most things you buy are going up faster than they did then, going   up as fast, going up slower, or not going up at all?                                                                                                                                                                                                                                                                                                                           \\
QQ9            & Do   you own a car?                                                                                                                                                                                                                                                                                                                                                                                                                                                                                    \\
QQ10           & In the past week, how many days did you   eat any butter?                                                                                                                                                                                                                                                                                                                                                                                                                                              \\
QQ11           & Since   your last treatment, discuss any personal problems if any at all that may be   related to your illness.                                                                                                                                                                                                                                                                                                                                                                                        \\
QQ12           & Do you use an answering machine?                                                                                                                                                                                                                                                                                                                                                                                                                                                                       \\
QQ13           & How   many peers do you have at work?                                                                                                                                                                                                                                                                                                                                                                                                                                                                  \\
QQ14           & What's your level of education?                                                                                                                                                                                                                                                                                                                                                                                                                                                                        \\
QQ15           & Do   you think it will be a problem if you become pregnant?                                                                                                                                                                                                                                                                                                                                                                                                                                            \\
QQ16           & Have you ever experienced a TIA or   micro-stroke?                                                                                                                                                                                                                                                                                                                                                                                                                                                     \\
QQ17           & Do   you think (your country) should implement a food stamps program?                                                                                                                                                                                                                                                                                                                                                                                                                                  \\
QQ18           & Do you support your congressman's recent   proposal to lower income taxes?                                                                                                                                                                                                                                                                                                                                                                                                                             \\
QQ19           & To   a tribal leader: tell the age of your kids.                                                                                                                                                                                                                                                                                                                                                                                                                                                       \\
QQ20           & Do you value your home?                                                                                                                                                                                                                                                                                                                                                                                                                                                                                \\
WVS1           & All   in all, how would you describe your state of health these days?                                                                                                                                                                                                                                                                                                                                                                                                                                  \\
WVS2           & All things considered, how satisfied are   you with your life as a whole these days?                                                                                                                                                                                                                                                                                                                                                                                                                   \\
WVS3           & How   satisfied are you with the financial situation of your household?                                                                                                                                                                                                                                                                                                                                                                                                                                \\
WVS4           & In the last 12 months, how often have   your or your family gone without enough food to eat?                                                                                                                                                                                                                                                                                                                                                                                                           \\
WVS5           & In   the last 12 months, how often have your or your family felt unsafe from crime   in your home?                                                                                                                                                                                                                                                                                                                                                                                                     \\
WVS6           & In the last 12 months, how often have   your or your family gone without medicine or medical treatment that you   needed?                                                                                                                                                                                                                                                                                                                                                                              \\
WVS7           & In   the last 12 months, how often have your or your family gone without a cash   income?                                                                                                                                                                                                                                                                                                                                                                                                              \\
WVS8           & In the last 12 months, how often have   your or your family gone without a safe shelter over your head?                                                                                                                                                                                                                                                                                                                                                                                                \\
WVS9           & Comparing   your standard of living with your parents’ standard of living when they were   about your age, would you say that you are better off, worse off or about the   same?                                                                                                                                                                                                                                                                                                                       \\
WVS10          & Generally speaking, would you say that   most people can be trusted or that you need to be very careful in dealing   with people?                                                                                                                                                                                                                                                                                                                                                                      \\
WVS11          & Nowadays   there’s a lot of talk about international organizations. People sometimes say   that international organizations should prioritize improving people’s lives,   even if this may imply that decisions are not made democratically. What do   you think should international organizations prioritize, being effective or   being democratic?                                                                                                                                                 \\
WVS12          & Five countries have permanent seats on   the Security Council of the United Nations. Which one of the following is not   a member: France, China, or India?                                                                                                                                                                                                                                                                                                                                            \\
WVS13          & Where   are the headquarters of the International Monetary Fund (IMF) located:   Washington DC, London, or Geneva?                                                                                                                                                                                                                                                                                                                                                                                     \\
WVS14          & Which of the following problems does the   organization Amnesty International deal with: Climate change, human rights,   or destruction of historic monuments?                                                                                                                                                                                                                                                                                                                                         \\
WVS15          & Now   I'd like you to tell me your views on corruption – when people pay a bribe,   give a gift or do a favor to other people in order to get the things they   need done or the services they need. How would you place your views on   corruption in your country on a 10-point scale where “1” means “there is no   corruption in my country” and “10” means “there is abundant corruption in my   country”.                                                                                        \\
WVS16          & We want to know about your experience   with local officials and service providers, like police officers, lawyers,   doctors, teachers and civil servants in your community. How often do you   think ordinary people like yourself or people from your neighbourhood have to   pay a bribe, give a gift or do a favor to these people in order to get the   services you need?                                                                                                                        \\
WVS17          & Can   you tell me how strongly you agree or disagree with the following statement:   “on the whole, women are less corrupt than men”?                                                                                                                                                                                                                                                                                                                                                                  \\
WVS18          & How high is the risk in this country to   be held accountable for giving or receiving a bribe, gift or favor in return   for public service?                                                                                                                                                                                                                                                                                                                                                           \\
WVS19          & Now   we would like to know your opinion about the people from other countries who   come to live in {[}your country{]} - the immigrants. How would you evaluate the   impact of these people on the development of {[}your country{]}?                                                                                                                                                                                                                                                                \\
WVS20          & How about people from other countries   coming here to work. Which one of the following do you think the government   should do?                                                                                                                                                                                                                                                                                                                                                                       \\
WVS21          & Could   you tell me how secure do you feel these days?                                                                                                                                                                                                                                                                                                                                                                                                                                                 \\
WVS22          & Have you been the victim of a crime   during the past year?                                                                                                                                                                                                                                                                                                                                                                                                                                            \\
WVS23          & And   what about your immediate family--has someone in your family been the victim   of a crime during the last year?                                                                                                                                                                                                                                                                                                                                                                                  \\
WVS24          & Most people consider both freedom and   equality to be important, but if you had to choose between them, which one   would you consider more important?                                                                                                                                                                                                                                                                                                                                                \\
WVS25          & Most   people consider both freedom and security to be important, but if you had to   choose between them, which one would you consider more important?                                                                                                                                                                                                                                                                                                                                                \\
WVS26          & Of course, we all hope that there will   not be another war, but if it were to come to that, would you be willing to   fight for your country?                                                                                                                                                                                                                                                                                                                                                         \\
WVS27          & Science   and technology are making our lives healthier, easier, and more comfortable.                                                                                                                                                                                                                                                                                                                                                                                                                 \\
WVS28          & Because of science and technology, there   will be more opportunities for the next generation.                                                                                                                                                                                                                                                                                                                                                                                                         \\
WVS29          & To   what extent do you agree with the following statement: "We depend too   much on science and not enough on faith."                                                                                                                                                                                                                                                                                                                                                                                 \\
WVS30          & To what extent do you agree with the   following statement: "One of the bad effects of science is that it   breaks down people’s ideas of right and wrong."                                                                                                                                                                                                                                                                                                                                            \\
WVS31          & To   what extent do you agree with the following statement: "It is not   important for me to know about science in my daily life."                                                                                                                                                                                                                                                                                                                                                                     \\
WVS32          & All things considered, would you say   that the world is better off, or worse off, because of science and   technology?                                                                                                                                                                                                                                                                                                                                                                                \\
WVS33          & How   important is God in your life?                                                                                                                                                                                                                                                                                                                                                                                                                                                                   \\
WVS34          & To what extent do you agree with the   following statement: "Whenever science and religion conflict, religion   is always right."                                                                                                                                                                                                                                                                                                                                                                      \\
WVS35          & To   what extent do you agree with the following statement: "The only   acceptable religion is my religion."                                                                                                                                                                                                                                                                                                                                                                                           \\
WVS36          & Apart from weddings and funerals, about   how often do you attend religious services these days?                                                                                                                                                                                                                                                                                                                                                                                                       \\
WVS37          & Apart   from weddings and funerals, about how often do you pray?                                                                                                                                                                                                                                                                                                                                                                                                                                       \\
WVS38          & How much do you agree or disagree with   the statement that nowadays one often has trouble deciding which moral rules   are the right ones to follow?                                                                                                                                                                                                                                                                                                                                                  \\
WVS39          & Do   you think the country's government should or should not have the right to do   the following keep people under video surveillance in public areas                                                                                                                                                                                                                                                                                                                                                 \\
WVS40          & Do you think the country's government   should or should not have the right to do the following monitor all emails   and other information exchanged on the internet                                                                                                                                                                                                                                                                                                                                   \\
WVS41          & Do   you think the country's government should or should not have the right to do   the following collect information about anyone living in this country without   their knowledge                                                                                                                                                                                                                                                                                                                    \\
WVS42          & How interested would you say you are in   politics?                                                                                                                                                                                                                                                                                                                                                                                                                                                    \\
WVS43          & When   you get together with your friends, would you say you discuss political   matters frequently, occasionally or never?                                                                                                                                                                                                                                                                                                                                                                            \\
WVS44          & If there were a national election   tomorrow, for which party on this list would you vote?                                                                                                                                                                                                                                                                                                                                                                                                             \\
WVS45          & Some   people think that having honest elections makes a lot of difference in their   lives; other people think that it doesn’t matter much. How important would   you say is having honest elections for you—very important, rather important,   not very important or not at all important?                                                                                                                                                                                                          \\
WVS46          & How much would you say the political   system in your country allows people like you to have a say in what the   government does?                                                                                                                                                                                                                                                                                                                                                                      \\
WVS47          & In   political matters, people talk of "the left" and "the   right." How would you place your views on this scale, generally   speaking?                                                                                                                                                                                                                                                                                                                                                               \\
WVS48          & How important is it for you to live in a   country that is governed democratically?                                                                                                                                                                                                                                                                                                                                                                                                                    \\
WVS49          & And   how democratically is this country being governed today?                                                                                                                                                                                                                                                                                                                                                                                                                                         \\
WVS50          & How satisfied are you with how the   political system is functioning in your country these days?                                                                                                                                                                                                                                                                                                                                                                                                       \\
WVS51          & How   much respect is there for individual human rights nowadays in this country?                                                                                                                                                                                                                                                                                                                                                                                                                      \\
WVS52          & How proud are you to be (country's   nationality)?                                                                                                                                                                                                                                                                                                                                                                                                                                                     \\
WVS53          & Could   you tell me whether you trust people from your family completely, somewhat,   not very much or not at all?                                                                                                                                                                                                                                                                                                                                                                                     \\
WVS54          & Could you tell me whether you trust   people from your neighborhood completely, somewhat, not very much or not at   all?                                                                                                                                                                                                                                                                                                                                                                               \\
WVS55          & Could   you tell me whether you trust people you know personally completely,   somewhat, not very much or not at all?                                                                                                                                                                                                                                                                                                                                                                                  \\
WVS56          & Could you tell me whether you trust   people you meet for the first time completely, somewhat, not very much or not   at all?                                                                                                                                                                                                                                                                                                                                                                          \\
WVS57          & Could   you tell me whether you trust people of another religion completely,   somewhat, not very much or not at all?                                                                                                                                                                                                                                                                                                                                                                                  \\
WVS58          & Could you tell me whether you trust   people of another nationality completely, somewhat, not very much or not at   all?                                                                                                                                                                                                                                                                                                                                                                               \\
WVS59          & Could   you tell me how much confidence you have in major companies?                                                                                                                                                                                                                                                                                                                                                                                                                                   \\
WVS60          & Could you tell me how much confidence   you have in The North American Treaty Association (NATO)?                                                                                                                                                                                                                                                                                                                                                                                                      \\
WVS61          & Could   you tell me how much confidence you have in The World Bank?                                                                                                                                                                                                                                                                                                                                                                                                                                    \\
WVS62          & Could you tell me how much confidence   you have in Political Parties?                                                                                                                                                                                                                                                                                                                                                                                                                                 \\
WVS63          & Could   you tell me how much confidence you have in the armed forces?                                                                                                                                                                                                                                                                                                                                                                                                                                  \\
WVS64          & Could you tell me how much confidence   you have in Labor Unions?                                                                                                                                                                                                                                                                                                                                                                                                                                      \\
WVS65          & Could   you tell me how much confidence you have in universities?                                                                                                                                                                                                                                                                                                                                                                                                                                      \\
WVS66          & Could you tell me how much confidence   you have in environmental organizations?                                                                                                                                                                                                                                                                                                                                                                                                                       \\
WVS67          & Could   you tell me how much confidence you have in The World Health Organization   (WHO)?                                                                                                                                                                                                                                                                                                                                                                                                             \\
WVS68          & Could you tell me how much confidence   you have in The World Trade Organization (WTO)?                                                                                                                                                                                                                                                                                                                                                                                                                \\
WVS69          & Could   you tell me how much confidence you have in the civil service?                                                                                                                                                                                                                                                                                                                                                                                                                                 \\
WVS70          & Could you tell me how much confidence   you have in Charitable or humanitarian organizations?                                                                                                                                                                                                                                                                                                                                                                                                          \\
WVS71          & Are   you an active member, an inactive member, or not a member of a church or   religious organization?                                                                                                                                                                                                                                                                                                                                                                                               \\
WVS72          & Are you an active member, an inactive   member, or not a member of a sport or recreational organization,   football/baseball/rugby team?                                                                                                                                                                                                                                                                                                                                                               \\
WVS73          & Are   you an active member, an inactive member, or not a member of an art, music,   or educational organization?                                                                                                                                                                                                                                                                                                                                                                                       \\
WVS74          & Are you an active member, an inactive   member, or not a member of a labor union?                                                                                                                                                                                                                                                                                                                                                                                                                      \\
WVS75          & Are   you an active member, an inactive member, or not a member of a political   party?                                                                                                                                                                                                                                                                                                                                                                                                                \\
WVS76          & Are you an active member, an inactive   member, or not a member of an environmental organization?                                                                                                                                                                                                                                                                                                                                                                                                      \\
WVS77          & Are   you an active member, an inactive member, or not a member of a professional   association?                                                                                                                                                                                                                                                                                                                                                                                                       \\
WVS78          & Are you an active member, an inactive   member, or not a member of a humanitarian or charitable organization?                                                                                                                                                                                                                                                                                                                                                                                          \\
WVS79          & Are   you an active member, an inactive member, or not a member of a consumer   organization?                                                                                                                                                                                                                                                                                                                                                                                                          \\
WVS80          & Are you an active member, an inactive   member, or not a member of a self-help group, mutual aid group?                                                                                                                                                                                                                                                                                                                                                                                                \\
WVS81          & Are   you an active member, an inactive member, or not a member of a women's group?                                                                                                                                                                                                                                                                                                                                                                                                                    \\
G1             & How satisfied are you with your company   as a workplace?                                                                                                                                                                                                                                                                                                                                                                                                                                              \\
G2             & Do   you know what is expected of you at work?                                                                                                                                                                                                                                                                                                                                                                                                                                                         \\
G3             & Do you have the materials and equipment   to do your work right?                                                                                                                                                                                                                                                                                                                                                                                                                                       \\
G4             & At   work, do you have the opportunity to do what you do best every day?                                                                                                                                                                                                                                                                                                                                                                                                                               \\
G5             & \begin{tabular}[c]{@{}l@{}}In the last seven days, have you   received recognition or praise for doing good\\      \\      work?\end{tabular}                                                                                                                                                                                                                                                                                                                                                          \\
G6             & Does   your supervisor, or someone at work, seem to care about you as a person?                                                                                                                                                                                                                                                                                                                                                                                                                        \\
G7             & Is there someone at work who encourages   your development?                                                                                                                                                                                                                                                                                                                                                                                                                                            \\
G8             & At   work, do your opinions seem to count?                                                                                                                                                                                                                                                                                                                                                                                                                                                             \\
G9             & Does the mission/purpose of your company   make you feel your job is important?                                                                                                                                                                                                                                                                                                                                                                                                                        \\
G10            & Are   your associates (fellow employees) committed to doing quality work?                                                                                                                                                                                                                                                                                                                                                                                                                              \\
G11            & Do you have a best friend at work?                                                                                                                                                                                                                                                                                                                                                                                                                                                                     \\
G12            & In   the last six months, has someone at work talked to you about your progress?                                                                                                                                                                                                                                                                                                                                                                                                                       \\
G13            & In the last year, have you had   opportunities to learn and grow?                                                                                                                                                                                                                                                                                                                                                                                                                                      \\
L1             & Please   tell me if you agree or disagree that it is proper and right that everyone   act according to  Your local council   member/member of county assembly decisions, even when others disagree with   the decisions ?                                                                                                                                                                                                                                                                              \\
L2             & How much influence do clan elders or   other traditional leaders currently have in governing your local community?                                                                                                                                                                                                                                                                                                                                                                                     \\
L3             & Would   you say that in general your Village Head/Neighborhood Block Leade is fair in   their treatment of members of this community?                                                                                                                                                                                                                                                                                                                                                                  \\
L4             & Since \{current month\} of last year, did   you or someone else in your household go without the necessary food,   clothing, and or shelter?                                                                                                                                                                                                                                                                                                                                                           \\
L5             & Did   you/someone else in your household turn to someone to help you access food,   clothing, or shelter?                                                                                                                                                                                                                                                                                                                                                                                              \\
L6             & Would you say that you used personal   connections in order to obtain access that you otherwise would not have   received, or to obtain faster/better service, or for some other reason?                                                                                                                                                                                                                                                                                                               \\
L7             & Did   you have to pay for this person for the assistance?                                                                                                                                                                                                                                                                                                                                                                                                                                              \\
L8             & How many of your children have a birth   certificate?                                                                                                                                                                                                                                                                                                                                                                                                                                                  \\
L9             & Did   you or anyone in your household attempt to obtain a birth certificate for   yourself or one of your children since \{current month\} of last year?                                                                                                                                                                                                                                                                                                                                               \\
L10            & If you have not attempted to obtain a   birth certificate why not?                                                                                                                                                                                                                                                                                                                                                                                                                                     \\
L11            & Ultimately,   did you obtain the business permit?                                                                                                                                                                                                                                                                                                                                                                                                                                                      \\
L12            & Did you have to pay extra money or give   a gift to get this service?                                                                                                                                                                                                                                                                                                                                                                                                                                  \\
L13            & Can   you tell me about an influential person in this village/neighborhood?                                                                                                                                                                                                                                                                                                                                                                                                                            \\
L14            & Who do you fear would physically punish   you?                                                                                                                                                                                                                                                                                                                                                                                                                                                         \\
L15            & Imagine   you are applying for a job, do you think that you have a better or worse   chance of being hired if the person making the hiring decisions is a   \{respondent's ethnic group\}?                                                                                                                                                                                                                                                                                                             \\
L16            & What is your religion, if any?                                                                                                                                                                                                                                                                                                                                                                                                                                                                         \\
L17            & How   often do you attend church, mosque or other religious services or meetings?                                                                                                                                                                                                                                                                                                                                                                                                                      \\
L18            & Have you heard of meetings, ceremonies   or other events that have been organized by women, specifically for women,   since \{current month\} of last year?                                                                                                                                                                                                                                                                                                                                            \\
L19            & Have   you heard of meetings, ceremonies or other events that have been organized by   men, specifically for men, since \{current month\} of last year?                                                                                                                                                                                                                                                                                                                                                \\
L20            & In your family, is a dowry paid when   people get married?                                                                                                                                                                                                                                                                                                                                                                                                                                             \\
L21            & Are   women more likely, less likely or neither more or less likely to have similar   opinions about important social issues (e.g., marriage and divorce, or   childrearing) with other women versus men?                                                                                                                                                                                                                                                                                              \\
L22            & Are women more obligated to help each   other, less obligated to help each other or neither more or less obligated to   help each other than they are to help men?                                                                                                                                                                                                                                                                                                                                     \\
L23            & In   general, are women more worried, less worried or equally worried about being   cheated when interacting with other women than they are when interacting with   men?                                                                                                                                                                                                                                                                                                                               \\
L24            & If you needed help from a government   worker getting access to a service such as public healthcare or enrollment of   your child into public school, do you think you get better assistance, worse   assistance, or the same assistance if the person is a woman?                                                                                                                                                                                                                                     \\
L25            & How   likely is it that a woman would be treated badly if she were to publicly   disagree with most other women on important issues?                                                                                                                                                                                                                                                                                                                                                                   \\
L26            & Imagine you are applying for a job, do   you think that you have a better or worse chance of being hired if the person   making the hiring decisions is a woman?                                                                                                                                                                                                                                                                                                                                       \\
L27            & Are   men more likely, less likely or neither more or less likely to have similar   opinions about important social issues (e.g., marriage and divorce, or   childrearing) with other women versus men?                                                                                                                                                                                                                                                                                                \\
L28            & Are men more obligated to help each   other, less obligated to help each other or neither more or less obligated to   help each other than they are to help men?                                                                                                                                                                                                                                                                                                                                       \\
L29            & In   general, are men more worried, less worried or equally worried about being   cheated when interacting with other women than they are when interacting with   men?                                                                                                                                                                                                                                                                                                                                 \\
L30            & If you needed help from a government   worker getting access to a service such as public healthcare or enrollment of   your child into public school, do you think you get better assistance, worse   assistance, or the same assistance if the person is a man?                                                                                                                                                                                                                                       \\
L31            & How   likely is it that a woman would be treated badly if she were to publicly   disagree with most other men on important issues?                                                                                                                                                                                                                                                                                                                                                                     \\
L32            & Imagine you are applying for a job, do   you think that you have a better or worse chance of being hired if the person   making the hiring decisions is a man?                                                                                                                                                                                                                                                                                                                                         \\
L33            & How   long have you lived in this village/neighborhood?                                                                                                                                                                                                                                                                                                                                                                                                                                                \\
L34            & How many adults (those aged 18 years and   above) presently live in this household? A household is a group of people who   presently eat together from the same pot.                                                                                                                                                                                                                                                                                                                                   \\
L35            & How   many children under the age of 18 live with you in this household?                                                                                                                                                                                                                                                                                                                                                                                                                               \\
L36            & How many people are enrolled in primary   or secondary school, whether living here or boarding elsewhere?                                                                                                                                                                                                                                                                                                                                                                                              \\
L37            & Do   you have a job that pays a cash income/salary?                                                                                                                                                                                                                                                                                                                                                                                                                                                    \\
L38            & Do you work in a non-cash-paying job, or   are you retired, disabled, not looking for work or searching for a job?                                                                                                                                                                                                                                                                                                                                                                                     \\
L39            & Do   you work for yourself, for someone else in the private sector or the   non-governmental sector, or for government?                                                                                                                                                                                                                                                                                                                                                                                \\
L40            & What is the main source of income for   your self-employment?                                                                                                                                                                                                                                                                                                                                                                                                                                          \\
L41            & What   is your status/relationship at home?                                                                                                                                                                                                                                                                                                                                                                                                                                                            \\
L42            & Do you or your household own/have the   following: Motor vehicle or motorcycle                                                                                                                                                                                                                                                                                                                                                                                                                         \\
L43            & Do   you or your household own/have the following: Mobile telephone                                                                                                                                                                                                                                                                                                                                                                                                                                    \\
L44            & Do you or your household own/have the   following: Radio                                                                                                                                                                                                                                                                                                                                                                                                                                               \\
L45            & Do   you or your household own/have the following: Smartphone                                                                                                                                                                                                                                                                                                                                                                                                                                          \\
L46            & Do you or your household own/have the   following: Television                                                                                                                                                                                                                                                                                                                                                                                                                                          \\
L47            & Do   you or your household own/have the following: Refrigerator                                                                                                                                                                                                                                                                                                                                                                                                                                        \\
L48            & Do you or your household own/have the   following: Bicycle                                                                                                                                                                                                                                                                                                                                                                                                                                             \\
L49            & Do   you or your household own/have the following: Bed                                                                                                                                                                                                                                                                                                                                                                                                                                                 \\
L50            & Do you or your household own/have the   following: Sofa Set                                                                                                                                                                                                                                                                                                                                                                                                                                            \\
L51            & What   is your wife's ethnic group?                                                                                                                                                                                                                                                                                                                                                                                                                                                                    \\
L52            & What is your husband's ethnic   group?                                                                                                                                                                                                                                                                                                                                                                                                                                                                 \\
L53            & What   is your father's ethnic group?                                                                                                                                                                                                                                                                                                                                                                                                                                                                  \\
L54            & What is your mother's ethnic group?                                                                                                                                                                                                                                                                                                                                                                                                                                                                    \\
L55            & Are   you a citizen of this country?                                                                                                                                                                                                                                                                                                                                                                                                                                                                   \\
L56            & Were you born in a rural village, a   small town, the capitol or another city?                                                                                                                                                                                                                                                                                                                                                                                                                         \\
L57            & In   which District/County is your birthplace located?                                                                                                                                                                                                                                                                                                                                                                                                                                                 \\
L58            & Did you apply for a driver's license   since \{current month\} last year?                                                                                                                                                                                                                                                                                                                                                                                                                              \\
L59            & What   is \{name of child\}'s age?                                                                                                                                                                                                                                                                                                                                                                                                                                                                     \\
L60            & What standard or form is \{name of child\}   in?                                                                                                                                                                                                                                                                                                                                                                                                                                                       \\
L61            & What   is the name of the school that \{name of child\} attends?                                                                                                                                                                                                                                                                                                                                                                                                                                       \\
L62            & Where is the school that \{name of child\}   attends located?                                                                                                                                                                                                                                                                                                                                                                                                                                          \\
L63            & Is   this school public or private?                                                                                                                                                                                                                                                                                                                                                                                                                                                                    \\
L64            & Is this school run by a faith-based   organization?                                                                                                                                                                                                                                                                                                                                                                                                                                                    \\
L65            & Is   there a parent-teacher organization (PTO/PTA) (or school management committee   depending on level of school) at any of your children/ your child's school?                                                                                                                                                                                                                                                                                                                                       \\
L66            & Are you a member of the PTA or school   management committee?                                                                                                                                                                                                                                                                                                                                                                                                                                          \\
L67            & Are   children from some villages or neighborhoods treated better than others?                                                                                                                                                                                                                                                                                                                                                                                                                         \\
L68            & Students receiving tutoring   treated better, worse or the same as those who do not?                                                                                                                                                                                                                                                                                                                                                                                                                   \\
L69            & Who   do you think would make them pay fines or make them suffer from material   losses?                                                                                                                                                                                                                                                                                                                                                                                                               \\
L70            & Are you eligible to vote in \{name of   country\}?                                                                                                                                                                                                                                                                                                                                                                                                                                                     \\
L71            & Do   you expect that an elected official would be more responsive or less   responsive to someone who Voted for them, over one who did not                                                                                                                                                                                                                                                                                                                                                             \\
L72            & Do you expect that a female   parliamentarian would be more responsive to women than men?                                                                                                                                                                                                                                                                                                                                                                                                              \\
L73            & Do   you expect that a male parliamentarian would be more responsive to men than   women?                                                                                                                                                                                                                                                                                                                                                                                                              \\
L74            & Do you think that an elected official   provides better services to a village or neighborhood that has unanimously   supported the candidate?                                                                                                                                                                                                                                                                                                                                                          \\
L75            & During   elections, do your friends and family organize to support the same candidate?                                                                                                                                                                                                                                                                                                                                                                                                                 \\
L76            & Where were you registered to vote during   the last elections?                                                                                                                                                                                                                                                                                                                                                                                                                                         \\
L77            & Do   you think that if a friend or someone from your family openly supported a   candidate that friends or family members did not like they would risk Losing   the enjoyment and company of others?                                                                                                                                                                                                                                                                                                   \\
L78            & Do you think that if a friend or someone   from your family openly supported a candidate that friends or family members   did not like they would risk Not hearing news and gaining information?                                                                                                                                                                                                                                                                                                       \\
L79            & Do   you think that if a friend or someone from your family openly supported a   candidate that friends or family members did not like they would risk Their   household not getting goodies and gifts?                                                                                                                                                                                                                                                                                                \\
L80            & Do you think that if a friend or someone   from your family openly supported a candidate that friends or family members   did not like they would risk Others thinking poorly of them or their   household?                                                                                                                                                                                                                                                                                            \\
L81            & Do   you think that if a friend or someone from your family openly supported a   candidate that friends or family members did not like they would risk Having   to pay fines or losing property?                                                                                                                                                                                                                                                                                                       \\
L82            & Do you think that if a friend or someone   from your family openly supported a candidate that friends or family members   did not like they would risk Being physically punished if they don't   cooperate?                                                                                                                                                                                                                                                                                            \\
L83            & During   elections, do most others in your village organize to support the same   candidate?                                                                                                                                                                                                                                                                                                                                                                                                           \\
L84            & Do you think that if someone from your   village/neighborhood openly supported a candidate that others in the   village/neighborhood did not like they would risk Losing the enjoyment and   company of others?                                                                                                                                                                                                                                                                                        \\
L85            & Do   you think that if someone from your village/neighborhood openly supported a   candidate that others in the village/neighborhood did not like they would   risk Not hearing news and gaining information?                                                                                                                                                                                                                                                                                          \\
L86            & Understanding that some people were   unable to vote in the last general elections, which of the following   statements is true for you?                                                                                                                                                                                                                                                                                                                                                               \\
L87            & During   elections, do women in your community organize to support the same candidate?                                                                                                                                                                                                                                                                                                                                                                                                                 \\
L88            & If the presidential elections were held   tomorrow, which party's candidate would you vote for?                                                                                                                                                                                                                                                                                                                                                                                                        \\
L89            & If   the parliamentary elections were held tomorrow, which party's candidate would   you vote for?                                                                                                                                                                                                                                                                                                                                                                                                     \\
L90            & During elections, do most \{respondent's   ethnicity\}'s organize to support the same candidate?                                                                                                                                                                                                                                                                                                                                                                                                       \\
L91            & Imagine   that you have a parliamentarian who has helped you or your family gain   personal benefits like a job, fertilizer subsidies, or housing assistance in   the past. How obligated do you feel to vote for this parliamentarian in the   elections?                                                                                                                                                                                                                                             \\
L92            & Please tell me if you think that it is   acceptable for religious leaders to try to influence others regarding   politics?                                                                                                                                                                                                                                                                                                                                                                             \\
L93            & If   the local elections were held tomorrow, which party's candidate would you   vote for?                                                                                                                                                                                                                                                                                                                                                                                                             \\
L94            & Please tell me if you think that it is   acceptable for village heads to try to influence others regarding politics?                                                                                                                                                                                                                                                                                                                                                                                   \\
L95            & Please   tell me if you think that it is acceptable for an ethnic spokesman to try to   influence others regarding politics?                                                                                                                                                                                                                                                                                                                                                                           \\
L96            & Do you think that others in your   community can know who you voted for in an election, if you do not tell them?                                                                                                                                                                                                                                                                                                                                                                                       \\
L97            & In   your area, are people like you intimidated to vote for certain candidates?                                                                                                                                                                                                                                                                                                                                                                                                                        \\
L98            & Did you or a member of your family   experience election related violence during the last election period?                                                                                                                                                                                                                                                                                                                                                                                             \\
L99            & Sometimes   politics and government can seem so complicated that a person like me can’t   really understand what is going on. Please tell me if you agree or disagree   with that statement.                                                                                                                                                                                                                                                                                                           \\
L100           & Do you think it is moral or immoral for   candidates to pass out gifts, money or personal favors during election time?                                                                                                                                                                                                                                                                                                                                                                                 \\
L101           & Since   \{current month\} of last year, has your house had access to electricity?                                                                                                                                                                                                                                                                                                                                                                                                                      \\
L102           & What is the source of electricity?                                                                                                                                                                                                                                                                                                                                                                                                                                                                     \\
L103           & Who   provides the electricity to your household ?                                                                                                                                                                                                                                                                                                                                                                                                                                                     \\
L104           & Does your household pay for electricity?                                                                                                                                                                                                                                                                                                                                                                                                                                                               \\
L105           & Do   you think you have a harder time than others in your community to access   electricity?                                                                                                                                                                                                                                                                                                                                                                                                           \\
L106           & Would you say that the government always   has the right to demand taxes from the citizens?                                                                                                                                                                                                                                                                                                                                                                                                            \\
L107           & In   general, do you think that the money you pay in taxes goes to the government   or to the pocket of those who collect the tax?                                                                                                                                                                                                                                                                                                                                                                     \\
L108           & Thinking of taxes, do you agree or   disagree that those who pay taxes benefit from this in the form of goods,   services and protection?                                                                                                                                                                                                                                                                                                                                                              \\
L109           & Were   the tax fees publicly announced or displayed?                                                                                                                                                                                                                                                                                                                                                                                                                                                   \\
L110           & Why did you not pay tax?                                                                                                                                                                                                                                                                                                                                                                                                                                                                               \\
L111           & Sometimes   people make contributions in order to obtain services such as access to land   or land titling, help resolving disputes, assistance in gaining government   documents. Since \{current month\} of last year, did you or anyone in your   household give money, labor or any other contribution to a village head,   traditional authority, armed group, religious leader, political party or   other community leader for any other reason than those we have talked about   so far today? \\
L112           & Since \{current month\} last year, have   you or any household member : been pregnant?                                                                                                                                                                                                                                                                                                                                                                                                                 \\
L113           & Since   \{current month\} last year, have you or any household member : delivered a   baby?                                                                                                                                                                                                                                                                                                                                                                                                            \\
L114           & Since \{current month\} last year, have   you or any household member : suffered an injury or been ill?                                                                                                                                                                                                                                                                                                                                                                                                \\
L115           & Thinking   about the last delivery, illness, or injury was this you or someone else in   the household?                                                                                                                                                                                                                                                                                                                                                                                                \\
L116           & What was the last health issue you or   someone in your household faced?                                                                                                                                                                                                                                                                                                                                                                                                                               \\
L117           & Does   the household own any land in this village/neighborhood or community?                                                                                                                                                                                                                                                                                                                                                                                                                           \\
L118           & How big is this land?                                                                                                                                                                                                                                                                                                                                                                                                                                                                                  \\
L119           & Is   this land used for residence, agricultural crops, gardening, livestock, a   small business or shop?                                                                                                                                                                                                                                                                                                                                                                                               \\
L120           & How many years have you used this piece   of land?                                                                                                                                                                                                                                                                                                                                                                                                                                                     \\
L121           & How   did this household acquire this land?                                                                                                                                                                                                                                                                                                                                                                                                                                                            \\
L122           & From whom did you acquire this land?                                                                                                                                                                                                                                                                                                                                                                                                                                                                   \\
L123           & In   this community, is land generally inherited through the man or the woman's   lineage or both?                                                                                                                                                                                                                                                                                                                                                                                                     \\
L124           & In general, who has more power to make   decisions about land in this community, women or men?                                                                                                                                                                                                                                                                                                                                                                                                         \\
L125           & Are   you worried that your land will be taken away from you?                                                                                                                                                                                                                                                                                                                                                                                                                                          \\
L126           & Who makes the primary decisions   regarding the use and management of the land?                                                                                                                                                                                                                                                                                                                                                                                                                        \\
L127           & Does   the household have a document that says s/he owns this land?                                                                                                                                                                                                                                                                                                                                                                                                                                    \\
L128           & Since \{current month\} of last year how   often have you: Sent money to family where you used to live?                                                                                                                                                                                                                                                                                                                                                                                                \\
L129           & Since   \{current month\} of last year how often have you: Received money or assistance   from family where you used to live?                                                                                                                                                                                                                                                                                                                                                                          \\
L130           & When you visit your relatives in \{name   of previous village\} do you bring gifts or other material goods with you for   them?                                                                                                                                                                                                                                                                                                                                                                        \\
L131           & How   long does it take you to travel from \{name of previous village\} to this   village?                                                                                                                                                                                                                                                                                                                                                                                                             \\
L132           & Did you turn to anyone for help settling   into this village/neighborhood?                                                                                                                                                                                                                                                                                                                                                                                                                             \\
L133           & Where   did you live before you moved here?                                                                                                                                                                                                                                                                                                                                                                                                                                                            \\
L134           & How old were you when you left \{name of   previous village\}?                                                                                                                                                                                                                                                                                                                                                                                                                                         \\
L135           & Why   did you choose to move to this village/neighborhood?                                                                                                                                                                                                                                                                                                                                                                                                                                             \\
L136           & How often do you or any member of your   household go to neighboring countries to visit friends or relatives?                                                                                                                                                                                                                                                                                                                                                                                          \\
L137           & Have   you ever considered moving permanently from here to another   village/neighborhood?                                                                                                                                                                                                                                                                                                                                                                                                             \\
L138           & Why did you consider moving?                                                                                                                                                                                                                                                                                                                                                                                                                                                                           \\
L139           & Where   did you consider moving?                                                                                                                                                                                                                                                                                                                                                                                                                                                                       \\
L140           & Do you have any land in \{name of   previous village\}?                                                                                                                                                                                                                                                                                                                                                                                                                                                \\
L141           & Why   did you stay in this village/neighborhood?                                                                                                                                                                                                                                                                                                                                                                                                                                                       \\
L142           & Where do you consider yourself to be   'from'?                                                                                                                                                                                                                                                                                                                                                                                                                                                         \\
L143           & Since   \{current month\} of last year, about how often have you Communicated with   someone living in that village?                                                                                                                                                                                                                                                                                                                                                                                   \\
L144           & Since \{current month\} of last year,   about how often have you Had someone from that village visit you?                                                                                                                                                                                                                                                                                                                                                                                              \\
L145           & Since   \{current month\} of last year, about how often have you Visited your home   village?                                                                                                                                                                                                                                                                                                                                                                                                          \\
L146           & Since \{current month\} of last year,   about how often have you Sent money to family members or others living in   that village?                                                                                                                                                                                                                                                                                                                                                                      \\
L147           & Since   \{current month\} of last year, about how often have you Received money or   assistance from someone in that village/neighborhood?                                                                                                                                                                                                                                                                                                                                                             \\
L148           & Do you own land or other significant   property in that area?                                                                                                                                                                                                                                                                                                                                                                                                                                          \\
L149           & Do   you plan to return to live in this area in the future?                                                                                                                                                                                                                                                                                                                                                                                                                                            \\
L150           & What relatives, if any, do you still   have living in \{name of previous village\}?                                                                                                                                                                                                                                                                                                                                                                                                                    \\
L151           & Since   \{current month\} of last year how often have you: Communicated with family   where you used to live?                                                                                                                                                                                                                                                                                                                                                                                          \\
L152           & Since \{current month\} of last year how   often have you: Visited family where you used to live?                                                                                                                                                                                                                                                                                                                                                                                                      \\
L153           & Have   you gotten married since \{current month\} last year?                                                                                                                                                                                                                                                                                                                                                                                                                                           \\
L154           & Did you apply for a marriage certificate   since \{current month\} last year?                                                                                                                                                                                                                                                                                                                                                                                                                          \\
L155           & Have   you wanted to obtain a passport since \{month when survey was taken\} last   year?                                                                                                                                                                                                                                                                                                                                                                                                              \\
L156           & Have you applied for a passport since   \{current month\}  last year?                                                                                                                                                                                                                                                                                                                                                                                                                                  \\
L157           & If   you have not attempted to obtain a passport why not?                                                                                                                                                                                                                                                                                                                                                                                                                                              \\
L158           & What is the name of this   village/neighborhood?                                                                                                                                                                                                                                                                                                                                                                                                                                                       \\
L159           & What   do you consider the three most serious problems facing the   village/neighborhood?                                                                                                                                                                                                                                                                                                                                                                                                              \\
L160           & When you think about your most immediate   neighbors, are most, some, only a few or none related to you?                                                                                                                                                                                                                                                                                                                                                                                               \\
L161           & How   often do you visit others in this village/neighborhood?                                                                                                                                                                                                                                                                                                                                                                                                                                          \\
L162           & Thinking about "outsiders,"   would you be willing to live next door to them?                                                                                                                                                                                                                                                                                                                                                                                                                          \\
L163           & Thinking   about "outsiders," would you be willing to allow your children to   marry them?                                                                                                                                                                                                                                                                                                                                                                                                             \\
L164           & Do you consider yourself to be an   outsider in this village or neighborhood?                                                                                                                                                                                                                                                                                                                                                                                                                          \\
L165           & What   is the main source of drinking water for members of your household?                                                                                                                                                                                                                                                                                                                                                                                                                             \\
L166           & Since \{current month\} of last year, how   often has your household gone without enough clean water for home use?                                                                                                                                                                                                                                                                                                                                                                                     \\
L167           & Does   your household pay to access water?                                                                                                                                                                                                                                                                                                                                                                                                                                                             \\
L168           & What happens (or what would happen) if   your household cannot pay for water?                                                                                                                                                                                                                                                                                                                                                                                                                       
\end{longtable}
 
\setcounter{table}{0}
\renewcommand{\thetable}{A\arabic{table}}
\setcounter{figure}{0}
\renewcommand{\thefigure}{A\arabic{figure}}

\clearpage

\section{Common Source Survey Language} \label{Appendix:commonsourcelang}
The following are words and phrases that would be coded as 1 in the qualitative coding of the AI output. They emerged and were accepted by both coauthors. Output coded as 1 (common source survey language) supersedes other codes because it would automatically flag issues like inconsistent conceptualization, formality, and syntax. The only exception is that 1 can sometimes be coded with 7 if the phrase exists but has specific regional terminology or coded with 8 if that type of question does not exist at all. 
\begin{itemize}
    \item A year ago
\item Agree/disagree
\item If any at all
\item Most important
\item Ranking 
\item scale used for agreement or disagreement
\item ``for some other reason”
\item Frequency: sometimes, always, daily, monthly, etc. 
\item ``like me”
\item How often
\item In general
\item Most, some, only, a few (as an answer set)
\item These days
\item More likely less likely  
\item Most people
\item To what extent
\item Have you heard of
\item Of course
\item Generally speaking
\item A person like me
\item All things considered
\item True for you
\item All in all

\end{itemize}

\setcounter{table}{0}
\renewcommand{\thetable}{A\arabic{table}}
\setcounter{figure}{0}
\renewcommand{\thefigure}{A\arabic{figure}}

\clearpage 

\section{Prompts}\label{Appendix:prompts}
\setcounter{table}{0}
\renewcommand{\thetable}{C\arabic{table}}
\setcounter{figure}{0}
\renewcommand{\thefigure}{C\arabic{figure}}

\subsection{Experiment prompts}

Give me up to 5 aspects of this question that will be difficult to translate into \{target and persona\}. ``\{question text\}"

\{Target and persona\}: translate into another language for people living in another country. You are an expert on survey translation, Castilian Spanish for people living in Spain. You are an expert on Spain, Mandarin Chinese for people living in Mainland China. You are an expert on mainland China

Note: Specification of Linguistic AND Cultural Context

\{question text\}: see Appendix A for complete experimental units

\textbf{Examples}

Give me up to 5 aspects of this question that will be difficult to translate into Castilian Spanish for people living in Spain. You are an expert on Spain. ``How many peers do you have at work?"

Give me up to 5 aspects of this question that will be difficult to translate into Mandarin Chinese for people living in Mainland China. You are an expert on mainland China. ``How many peers do you have at work?"

Give me up to 5 aspects of this question that will be difficult to translate into another language for people living in another country. You are an expert on survey translation. ``How many peers do you have at work?"

\subsection{Formatting Prompt}
Format your response as CSV file code with the first column named Treatment with the value {unique identifier}, second column named StatementNum equal to the number of your response, third column named Aspect equal to the aspect, and a fourth column named Description with the full description of the aspect verbatim from your response. 

\{unique identifier\}: E\{experiment number\}M\{model number\}T\{target linguistic\}\{question label\}

\{experiment number\}: 1=experiment 1 (not relevant for this study); 2=experiment 2

\{model number\}: 1=GPT3.5; 2=GPT4

\{target linguistic\}: 1=none/translation; 2=Castilian Spanish; 3=Mandarin Chinese

\{question label\}: see Appendix A for complete experimental units and corresponding label

\clearpage  

\section{Power Analysis}\label{Appendix:poweranalysis}
\setcounter{table}{0}
\renewcommand{\thetable}{E\arabic{table}}
\setcounter{figure}{0}
\renewcommand{\thefigure}{E\arabic{figure}}

\begin{figure}[htbp]
    \centering
    \caption{Power Analysis for Model Effect (logistic 2-level model)}
    \includegraphics[width=0.9\linewidth]{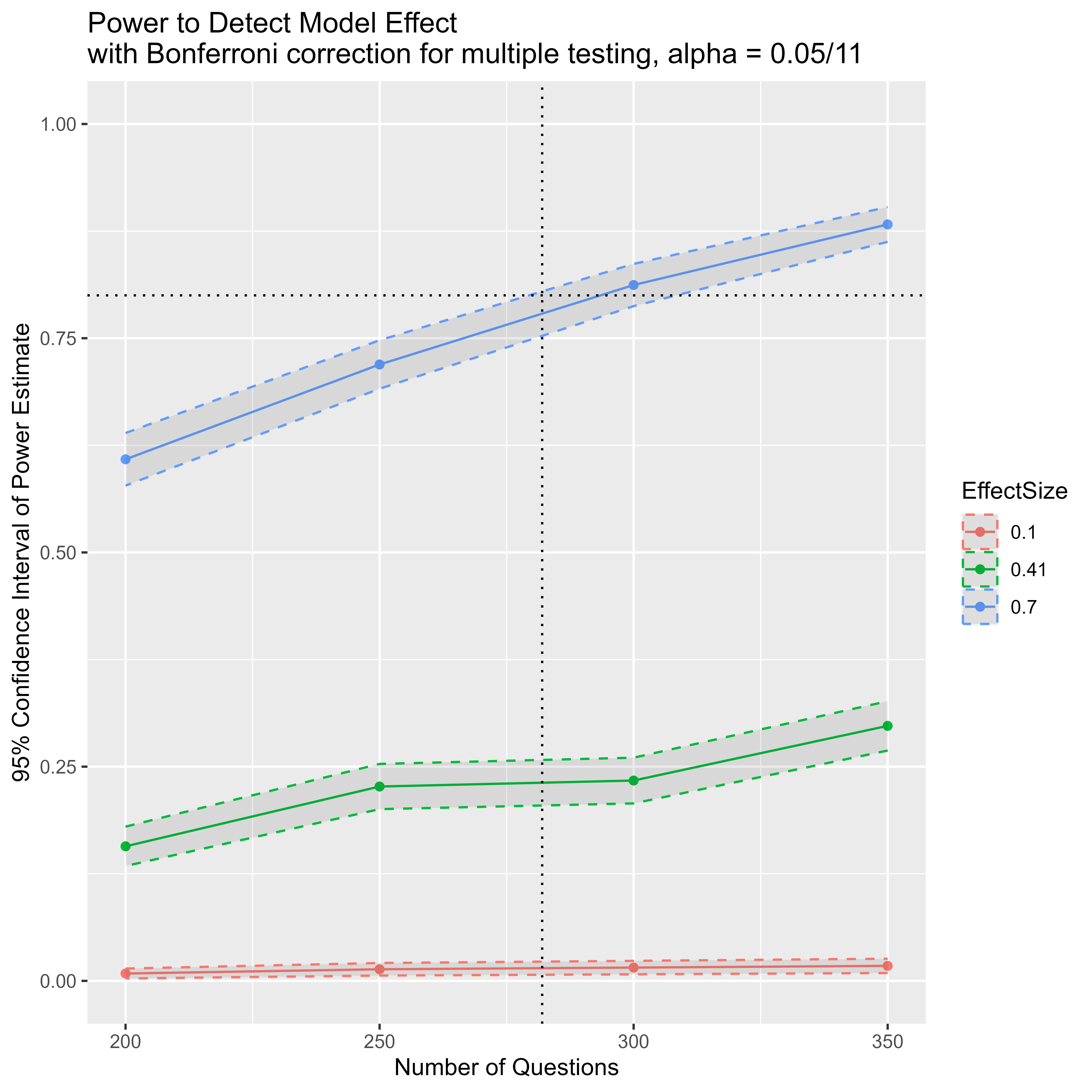}
    \label{fig:ModelPowerAnalysis}
\end{figure}

\begin{figure}[htbp]
    \centering
    \caption{Power Analysis for Persona (Target Linguistic Audience) Effect (logistic 2-level model)}
    \includegraphics[width=0.9\linewidth]{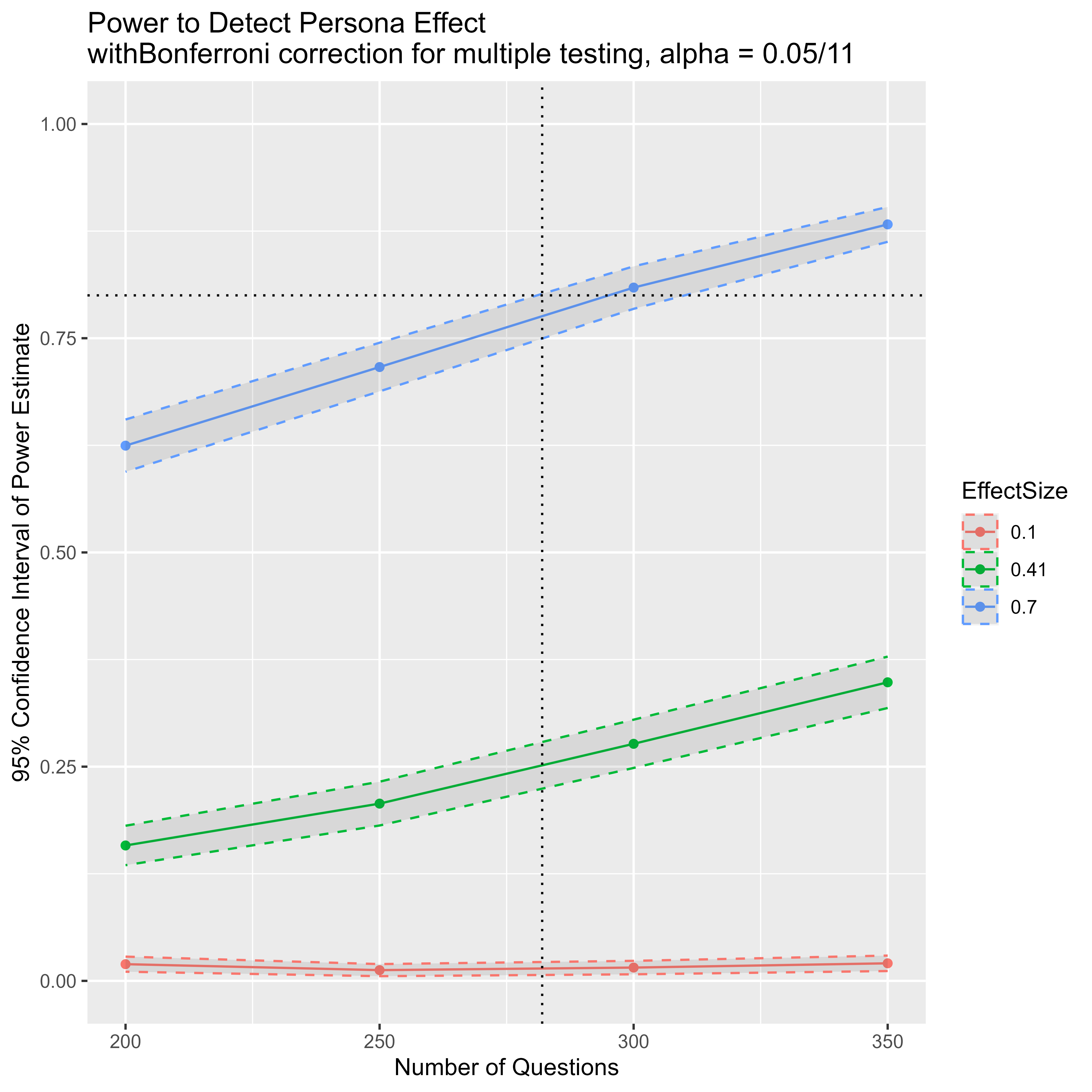}
    \label{fig:PersonaPowerAnalysis}
\end{figure}

\begin{figure}[htbp]
    \centering
    \caption{Power Analysis for Interaction Effect Between Model and Persona (logistic 2-level model)}
    \includegraphics[width=0.9\linewidth]{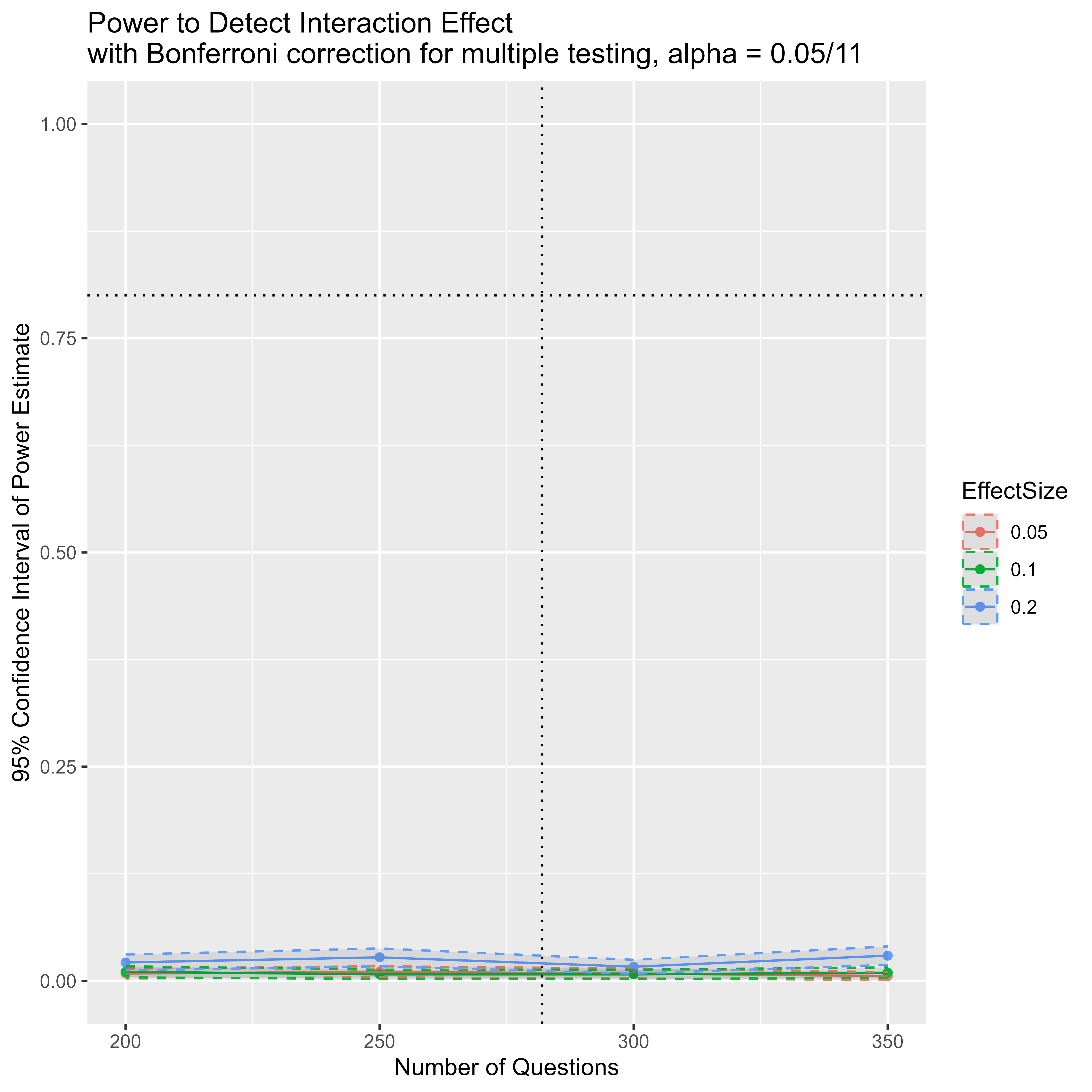}
    \label{fig:InteractionPowerAnalysis}
\end{figure}

\clearpage

\section{Qualitative Coding Details}\label{appendix:qualcodingdetails}

Qualitative coding was done in two different stages. In the first round, the output from the questionable questions were coded during codebook refinement, and the authors knew the source though not the treatment or question. Forr the first 100 statements, the authors coded together to decide a code and settle on the proper line of reasoning. Then in two batches of 100, the authors coded independently and settled discrepancies while updating the code book. With the final 300 questionable question statements, the authors coded independently and passed an inter-coder reliability check \footnote{The coders were above 80\% agreement for all codes and 70\% full agreement on statements.}. 

For the second round, the output from the other sources were randomized and anonymized at the statement level in an attempt to blind the qualitative coders from the source, treatment, and question.\footnote{Ultimately, coders were often able guess the target linguistic audience of Chinese/Spanish/Unspecified from the output.} The authors coded the first 100 together to ensure the codebook's applicability for output from a different question sources. Once this was clarified, they coded the remainder of the output independently and highlighted and decided codes for material that was unclear or novel.

\begin{landscape}

\begin{longtable}[t]{|p{6em}|p{10em}|p{15em}|p{7em}|p{12em}|}
\caption{Qualitative Codebook\label{tab:codescheme2}}\\
\hline
\multicolumn{1}{|c|}{\textbf{Code}} & \multicolumn{1}{|c|}{\textbf{Label}} &        \textbf{Definition} & \multicolumn{1}{|c|}{\textbf{Source}}                                           & \multicolumn{1}{|c|}{\textbf{Example}}                                                         \\ \hline
1    & Common source survey language   & Words/Phrases commonly used in L1surveys that are cumbersome or strange to translate& Harkness et al (2004)  & ``if any at all," see Appendix \ref{Appendix:commonsourcelang} for all adopted terms and language\\ \hline
2                               & Technical Terminology    & Specialized terms & Weeks et al (2007)    & medical terms, democracy   \\ \hline
3                               & Inconsistent conceptualization    & When the concept is not universal, if it’s translatable but the answer will be unclear, may include internal inconsistency within the country/culture & Smith (2004)  & Age, education, family    \\ \hline
4                               & Gendered Language   & When L1 uses a gendered word and translators won’t know whether to keep the gender, when L2 has gender translation issues & Harkness et al (2004)   & Chairman, actress  \\ \hline
5                               & Formality           & When the tone, politeness, or register may be relevant to the specific question & Harkness et al (2004)    & You/Thou   \\ \hline
6                               & Syntax & If you can translate but it won’t be smooth in L2, awkwardness and clunky, problems affiliated with tenses & Weeks et al (2007)       & Since your last treatment, discuss any personal problems that may be related to your illness. \\ \hline
7                               & Cultural/Regional Specific Terms    & Need to identify a word or phrase that exists in L1 and a different, specific version exists in L2, also regional variation & Harkness and Schoua-Glusberg (1998) & Parliament    \\ \hline
8                               & Non-existent concept  & Topic or concepts that do not meaningfully exist or matter in L2 & Weeks et al (2007)   & Food stamps      \\ \hline

9                              & Sensitive Topics    & Topic is emotional or taboo, may spark strong emotions & Smith (2004)   & God/Religion, pregnancy     \\ \hline
 Excluded in Analysis & Source language problem & Problems affiliated with the question irrespective of translation (not affiliated with multilinguistic or multicultural) & Rothgeb et al (2007) & Double barreled questions, scope issues, unfair presumption\\ \hline 
 NOTA & None of the above & Non-specific advice, piping issues, does not pertain to the question, nonsense or hallucination & n/a & n/a\\ \hline
\end{longtable}

\end{landscape}

\clearpage 

\section{Full Results}\label{Appendix:fullresults}
\setcounter{table}{0}
\renewcommand{\thetable}{F\arabic{table}}
\setcounter{figure}{0}
\renewcommand{\thefigure}{F\arabic{figure}}

\subsection{Data Summary}

\begin{figure}[htbp]
    \centering
    \caption{Distribution of Codes at the Statement Level}
    \includegraphics[width = 0.8\linewidth]{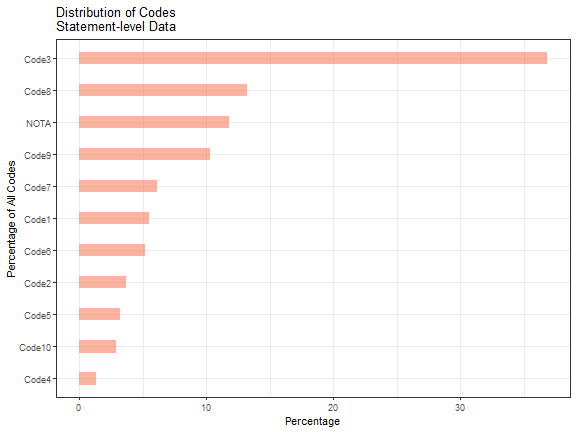}
    \label{fig:DistCode_Statement}
\end{figure}

\begin{figure}[htbp]
    \centering
    \caption{Distribution of Codes at the Treatment-Question Level}
    \includegraphics[width = 0.8\linewidth]{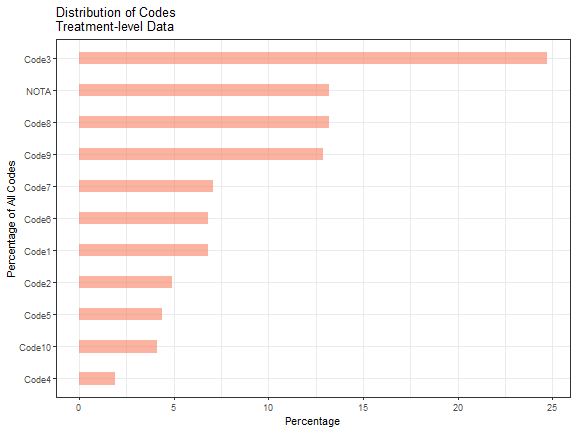}
    \label{fig:DistCode_TrtQ}
\end{figure}

\clearpage

\subsection{Regression Results}\

\subsubsection{Main Effects}

\begin{table}[htbp]
\begin{center}
\caption{Multilevel Regression Results for Number of Codes Per Treatment-Question Pair}
\label{tab:NumCodesTrtQ}
\begin{tabular}{l c}
\hline
 & Number of Codes \\
\hline
(Intercept)               & $3.24^{***}$ \\
                          & $(0.05)$     \\
ModelM2                   & $-0.10^{*}$  \\
                          & $(0.04)$     \\
AudienceT2                & $-0.08$      \\
                          & $(0.05)$     \\
AudienceT3                & $0.16^{**}$  \\
                          & $(0.05)$     \\
\hline
AIC                       & $4631.53$    \\
BIC                       & $4664.13$    \\
Log Likelihood            & $-2309.77$   \\
Num. obs.                 & $1692$       \\
Num. groups: Question     & $282$        \\
Var: Question (Intercept) & $0.14$       \\
Var: Residual             & $0.79$       \\
\hline
\multicolumn{2}{l}{\scriptsize{$^{***}p<0.001$; $^{**}p<0.01$; $^{*}p<0.05$}}
\end{tabular}

\end{center}
\end{table}

\begin{table}[htbp]
\begin{center}
\caption{Multilevel Regression Results the Code NOTA (None of the Above)}
\label{tab:RegNOTA}
\begin{tabular}{l c}
\hline
 & NOTA \\
\hline
(Intercept)               & $0.34^{***}$ \\
                          & $(0.02)$     \\
ModelM2                   & $-0.03$      \\
                          & $(0.02)$     \\
AudienceT2                & $0.32^{***}$ \\
                          & $(0.03)$     \\
AudienceT3                & $0.16^{***}$ \\
                          & $(0.03)$     \\
\hline
AIC                       & $2350.20$    \\
BIC                       & $2382.80$    \\
Log Likelihood            & $-1169.10$   \\
Num. obs.                 & $1692$       \\
Num. groups: Question     & $282$        \\
Var: Question (Intercept) & $0.02$       \\
Var: Residual             & $0.22$       \\
\hline
\multicolumn{2}{l}{\scriptsize{$^{***}p<0.001$; $^{**}p<0.01$; $^{*}p<0.05$}}
\end{tabular}

\end{center}
\end{table}

\clearpage

\subsubsection{Question Source}

\begin{table}[htbp]
\caption{Regression Results of Question Source Analysis}
\begin{center}
\begin{tabular}{l c}
\hline
 & Number of Codes \\
\hline
(Intercept)  & $3.41^{***}$  \\
             & $(0.10)$      \\
ModelM2      & $-0.10^{*}$   \\
             & $(0.05)$      \\
AudienceT2   & $-0.08$       \\
             & $(0.06)$      \\
AudienceT3   & $0.16^{**}$   \\
             & $(0.06)$      \\
SourceLGPI   & $-0.13$       \\
             & $(0.09)$      \\
SourceWVS    & $-0.24^{*}$   \\
             & $(0.10)$      \\
SourceGallup & $-0.61^{***}$ \\
             & $(0.14)$      \\
\hline
R$^2$        & $0.03$        \\
Adj. R$^2$   & $0.02$        \\
Num. obs.    & $1692$        \\
\hline
\multicolumn{2}{l}{\scriptsize{$^{***}p<0.001$; $^{**}p<0.01$; $^{*}p<0.05$}}
\end{tabular}
\label{reg:Source}
\end{center}
\end{table}

\begin{table}[htbp]
\caption{Multilevel Regression Results of Question Source Analysis}
\begin{center}
\begin{tabular}{l c}
\hline
 & Number of Codes \\
\hline
(Intercept)                      & $3.19^{***}$ \\
                                 & $(0.12)$     \\
ModelM2                          & $-0.10^{*}$  \\
                                 & $(0.04)$     \\
AudienceT2                       & $-0.08$      \\
                                 & $(0.05)$     \\
AudienceT3                       & $0.16^{**}$  \\
                                 & $(0.05)$     \\
\hline
AIC                              & $4628.88$    \\
BIC                              & $4666.92$    \\
Log Likelihood                   & $-2307.44$   \\
Num. obs.                        & $1692$       \\
Num. groups: Question:Source     & $282$        \\
Num. groups: Source              & $4$          \\
Var: Question:Source (Intercept) & $0.13$       \\
Var: Source (Intercept)          & $0.04$       \\
Var: Residual                    & $0.79$       \\
\hline
\multicolumn{2}{l}{\scriptsize{$^{***}p<0.001$; $^{**}p<0.01$; $^{*}p<0.05$}}
\end{tabular}
\label{reg:3MLMSource}
\end{center}
\end{table}

\clearpage 

\subsubsection{Interaction Effects}\label{appendix:interactions}

\begin{table}[htbp]
\begin{center}
\caption{Multilevel Regression Results with Model by Target Linguistic Audience Interaction - Number of Codes per Treatment-Question Pair}
\label{tab:RegNumCodesInt}
\begin{tabular}{l c}
\hline
 & Number of Codes \\
\hline
(Intercept)               & $3.15^{***}$ \\
                          & $(0.06)$     \\
ModelM2                   & $0.08$       \\
                          & $(0.07)$     \\
AudienceT2                & $0.09$       \\
                          & $(0.07)$     \\
AudienceT3                & $0.27^{***}$ \\
                          & $(0.07)$     \\
ModelM2:AudienceT2        & $-0.34^{**}$ \\
                          & $(0.11)$     \\
ModelM2:AudienceT3        & $-0.21^{*}$  \\
                          & $(0.11)$     \\
\hline
AIC                       & $4630.75$    \\
BIC                       & $4674.22$    \\
Log Likelihood            & $-2307.37$   \\
Num. obs.                 & $1692$       \\
Num. groups: Question     & $282$        \\
Var: Question (Intercept) & $0.14$       \\
Var: Residual             & $0.79$       \\
\hline
\multicolumn{2}{l}{\scriptsize{$^{***}p<0.001$; $^{**}p<0.01$; $^{*}p<0.05$}}
\end{tabular}

\end{center}
\end{table}

\begin{figure}[htbp]
    \centering
    \caption{Interaction Plot for the Interaction of Model and Target Audience on the Number of Codes Per Treatment-    \includegraphics[width = 0.8\linewidth]{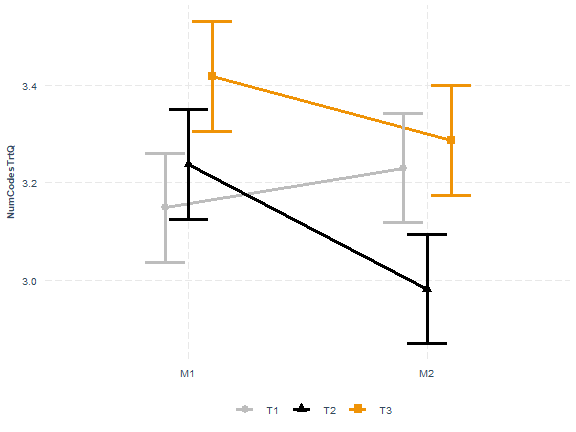}
Question Pair}
    \label{fig:NumCodesIntplot}
\end{figure}

\begin{table}
\caption{Multilevel Regression Results for Codes 1 to 5 - Model by Target Linguistic Audience Interaction}
\begin{center}
\begin{tabular}{l c c c c c}
\hline
 & Code 1 & Code 2 & Code 3 & Code 4 & Code 5 \\
\hline
(Intercept)               & $0.29^{***}$  & $0.19^{***}$ & $0.96^{***}$  & $0.07^{***}$ & $0.03$        \\
                          & $(0.03)$      & $(0.02)$     & $(0.02)$      & $(0.02)$     & $(0.02)$      \\
ModelM2                   & $0.01$        & $-0.05^{*}$  & $0.00$        & $-0.02$      & $0.07^{*}$    \\
                          & $(0.02)$      & $(0.03)$     & $(0.02)$      & $(0.02)$     & $(0.03)$      \\
AudienceT2                & $-0.05^{*}$   & $0.07^{**}$  & $-0.09^{***}$ & $0.01$       & $0.18^{***}$  \\
                          & $(0.02)$      & $(0.03)$     & $(0.02)$      & $(0.02)$     & $(0.03)$      \\
AudienceT3                & $-0.08^{***}$ & $0.02$       & $-0.06^{**}$  & $-0.01$      & $0.24^{***}$  \\
\end{tabular}
\label{tab:RegCode1to5.int}
\end{center}
\end{table}

\begin{table}
\caption{Multilevel Regression Results for Codes 6 to 10 - Model by Target Linguistic Audience Interaction}
\begin{center}
\begin{tabular}{l c c c c c}
\hline
 & Code 6 & Code 7 & Code 8 & Code 9 & Code 10 \\
\hline
(Intercept)               & $0.21^{***}$ & $0.20^{***}$  & $0.67^{***}$  & $0.36^{***}$ & $0.17^{***}$ \\
                          & $(0.03)$     & $(0.03)$      & $(0.03)$      & $(0.03)$     & $(0.02)$     \\
ModelM2                   & $0.06$       & $-0.03$       & $-0.00$       & $0.05$       & $-0.01$      \\
                          & $(0.03)$     & $(0.03)$      & $(0.04)$      & $(0.04)$     & $(0.03)$     \\
AudienceT2                & $0.05$       & $0.28^{***}$  & $-0.34^{***}$ & $0.01$       & $-0.04$      \\
                          & $(0.03)$     & $(0.03)$      & $(0.04)$      & $(0.04)$     & $(0.03)$     \\
AudienceT3                & $0.01$       & $0.08^{*}$    & $-0.17^{***}$ & $0.27^{***}$ & $-0.02$      \\
\end{tabular}
\label{tab:RegCode6to10.int}
\end{center}
\end{table}

\begin{table}
\caption{Multilevel Regression Results the Code NOTA (None of the Above) - Model by Target Linguistic Audience Interaction}
\begin{center}
\begin{tabular}{l c}
\hline
 & NOTA \\
\hline
(Intercept)               & $0.41^{***}$  \\
                          & $(0.03)$      \\
ModelM2                   & $-0.16^{***}$ \\
                          & $(0.04)$      \\
AudienceT2                & $0.22^{***}$  \\
                          & $(0.04)$      \\
AudienceT3                & $0.07$        \\
                          & $(0.04)$      \\
ModelM2:AudienceT2        & $0.21^{***}$  \\
                          & $(0.06)$      \\
ModelM2:AudienceT3        & $0.18^{***}$  \\
                          & $(0.06)$      \\
\hline
AIC                       & $2344.92$     \\
BIC                       & $2388.39$     \\
Log Likelihood            & $-1164.46$    \\
Num. obs.                 & $1692$        \\
Num. groups: Question     & $282$         \\
Var: Question (Intercept) & $0.02$        \\
Var: Residual             & $0.21$        \\
\hline
\multicolumn{2}{l}{\scriptsize{$^{***}p<0.001$; $^{**}p<0.01$; $^{*}p<0.05$}}
\end{tabular}
\label{tab:RegNOTA.int}
\end{center}
\end{table}

\clearpage

\subsubsection{Statement Order}

\begin{table}[htbp]
\begin{center}
\begin{threeparttable}
\caption{One--way ANOVA Test of Statement Placement by Code}\label{tab:ANOVA}
\begin{tabular}{lllllll}
\toprule
Effect & \multicolumn{1}{c}{$\hat{\eta}^2_G$} & \multicolumn{1}{c}{90\% CI} & \multicolumn{1}{c}{$F$} & \multicolumn{1}{c}{$\mathit{df}$} & \multicolumn{1}{c}{$\mathit{df}_{\mathrm{res}}$} & \multicolumn{1}{c}{$p$}\\
\midrule
Code & .077 & {}[.068, .085] & 83.68 & 9 & 9052 & $<$ .001\\
\bottomrule
\end{tabular}
\end{threeparttable}
\end{center}
\end{table}

\begin{longtable}[t]{>{}lrrrr}
\caption{Tukey HSD Test Results of Statement Placement by Code\label{tab:TukeyHSD}}\\
\toprule
 & Difference & Lower CI & Upper CI & Adjusted p-Value\\
\midrule
\cellcolor[HTML]{C1BBB5}{Code2-Code1} & 0.0590793 & -0.2413424 & 0.3595009 & 0.9998180\\
\cellcolor[HTML]{C1BBB5}{Code3-Code1} & -0.0312665 & -0.2357665 & 0.1732335 & 0.9999784\\
\cellcolor[HTML]{C1BBB5}{Code4-Code1} & -0.0848739 & -0.5236163 & 0.3538684 & 0.9998413\\
\cellcolor[HTML]{79C9C6}{Code5-Code1} & 0.6186633 & 0.3054177 & 0.9319088 & 0.0000000\\
\cellcolor[HTML]{79C9C6}{Code6-Code1} & 0.6694747 & 0.3962819 & 0.9426675 & 0.0000000\\
\addlinespace
\cellcolor[HTML]{C1BBB5}{Code7-Code1} & 0.1761185 & -0.0867829 & 0.4390199 & 0.5132596\\
\bottomrule
\end{longtable}

\clearpage

\subsection{Robustness Tests}\label{Appendix:robustness}

\subsubsection{Multilevel Logistic Models - Main Effects}

\begin{table}[htbp]
\begin{center}
\caption{Multilevel Logistic Regression Results for Codes 1 to 5}
\label{tab:RegCode1to5log}
\begin{tabular}{l c c c c c}
\hline
 & Code 1 & Code 2 & Code 3 & Code 4 & Code 5 \\
\hline
(Intercept)               & $-2.76^{***}$ & $-2.43^{***}$ & $4.53^{***}$  & $-8.90^{***}$ & $-3.21^{***}$ \\
                          & $(0.50)$      & $(0.25)$      & $(0.39)$      & $(0.78)$      & $(0.23)$      \\
ModelM2                   & $0.02$        & $-0.89^{***}$ & $0.18$        & $-0.49$       & $-0.16$       \\
                          & $(0.18)$      & $(0.17)$      & $(0.20)$      & $(0.30)$      & $(0.15)$      \\
AudienceT2                & $-0.67^{**}$  & $0.46^{*}$    & $-1.54^{***}$ & $1.24^{***}$  & $1.71^{***}$  \\
                          & $(0.21)$      & $(0.20)$      & $(0.29)$      & $(0.37)$      & $(0.22)$      \\
AudienceT3                & $-1.17^{***}$ & $0.06$        & $-1.25^{***}$ & $-0.59$       & $1.53^{***}$  \\
                          & $(0.23)$      & $(0.20)$      & $(0.30)$      & $(0.39)$      & $(0.22)$      \\
\hline
AIC                       & $1274.68$     & $1314.41$     & $864.91$      & $535.97$      & $1372.83$     \\
BIC                       & $1301.85$     & $1341.58$     & $892.08$      & $563.13$      & $1400.00$     \\
Log Likelihood            & $-632.34$     & $-652.21$     & $-427.46$     & $-262.98$     & $-681.42$     \\
Num. obs.                 & $1692$        & $1692$        & $1692$        & $1692$        & $1692$        \\
Num. groups: Question     & $282$         & $282$         & $282$         & $282$         & $282$         \\
Var: Question (Intercept) & $15.96$       & $4.92$        & $3.39$        & $57.66$       & $1.39$        \\
\hline
\multicolumn{6}{l}{\scriptsize{$^{***}p<0.001$; $^{**}p<0.01$; $^{*}p<0.05$}}
\end{tabular}
\end{center}
\end{table}

\begin{table}[htbp]
\label{tab:RegCode6to10log}
\caption{Multilevel Logistic Regression Results for Codes 6 to 10}
\begin{center}
\begin{tabular}{l c c c c c}
\hline
 & Code 6 & Code 7 & Code 8 & Code 9 & Code 10 \\
\hline
(Intercept)               & $-1.52^{***}$ & $-1.67^{***}$ & $0.95^{***}$  & $-0.77^{***}$ & $-1.91^{***}$ \\
                          & $(0.14)$      & $(0.17)$      & $(0.00)$      & $(0.13)$      & $(0.16)$      \\
ModelM2                   & $0.30^{*}$    & $-0.73^{***}$ & $-0.08^{***}$ & $0.32^{**}$   & $0.11$        \\
                          & $(0.12)$      & $(0.13)$      & $(0.00)$      & $(0.12)$      & $(0.14)$      \\
AudienceT2                & $0.28$        & $1.30^{***}$  & $-1.82^{***}$ & $-0.24$       & $-0.31$       \\
                          & $(0.15)$      & $(0.17)$      & $(0.00)$      & $(0.14)$      & $(0.18)$      \\
AudienceT3                & $-0.09$       & $0.47^{**}$   & $-1.03^{***}$ & $1.78^{***}$  & $-0.01$       \\
                          & $(0.15)$      & $(0.17)$      & $(0.00)$      & $(0.15)$      & $(0.17)$      \\
\hline
AIC                       & $1845.53$     & $1751.72$     & $2111.23$     & $2044.23$     & $1417.41$     \\
BIC                       & $1872.70$     & $1778.89$     & $2138.40$     & $2071.39$     & $1444.58$     \\
Log Likelihood            & $-917.76$     & $-870.86$     & $-1050.61$    & $-1017.11$    & $-703.70$     \\
Num. obs.                 & $1692$        & $1692$        & $1692$        & $1692$        & $1692$        \\
Num. groups: Question     & $282$         & $282$         & $282$         & $282$         & $282$         \\
Var: Question (Intercept) & $0.93$        & $1.83$        & $1.27$        & $1.28$        & $0.68$        \\
\hline
\multicolumn{6}{l}{\scriptsize{$^{***}p<0.001$; $^{**}p<0.01$; $^{*}p<0.05$}}
\end{tabular}

\end{center}
\end{table}

\begin{table}[htbp]
\begin{center}
\caption{Multilevel Logistic Regression Results the Code NOTA (None of the Above)}
\label{tab:RegNOTAlog}
\begin{tabular}{l c}
\hline
 & NOTA \\
\hline
(Intercept)               & $-0.71^{***}$ \\
                          & $(0.11)$      \\
ModelM2                   & $-0.13$       \\
                          & $(0.10)$      \\
AudienceT2                & $1.44^{***}$  \\
                          & $(0.13)$      \\
AudienceT3                & $0.73^{***}$  \\
                          & $(0.13)$      \\
\hline
AIC                       & $2218.17$     \\
BIC                       & $2245.34$     \\
Log Likelihood            & $-1104.08$    \\
Num. obs.                 & $1692$        \\
Num. groups: Question     & $282$         \\
Var: Question (Intercept) & $0.31$        \\
\hline
\multicolumn{2}{l}{\scriptsize{$^{***}p<0.001$; $^{**}p<0.01$; $^{*}p<0.05$}}
\end{tabular}

\end{center}
\end{table}

\clearpage

\subsubsection{Multilevel Logistic Models - Interaction Effects}

\begin{table}[htbp]
\caption{Multilevel Logistic Regression Results for Codes 1 to 5 - Model by Target Linguistic Audience Interaction}
\begin{center}
\begin{tabular}{l c c c c c}
\hline
 & Code 1 & Code 2 & Code 3 & Code 4 & Code 5 \\
\hline
(Intercept)               & $-2.82^{***}$ & $-2.55^{***}$ & $4.56^{***}$  & $-8.78^{***}$ & $-4.23^{***}$ \\
                          & $(0.51)$      & $(0.28)$      & $(0.44)$      & $(0.80)$      & $(0.40)$      \\
ModelM2                   & $0.13$        & $-0.62^{*}$   & $0.12$        & $-0.99$       & $1.41^{***}$  \\
                          & $(0.29)$      & $(0.29)$      & $(0.49)$      & $(0.54)$      & $(0.43)$      \\
AudienceT2                & $-0.58$       & $0.64^{*}$    & $-1.61^{***}$ & $0.50$        & $2.53^{***}$  \\
                          & $(0.30)$      & $(0.26)$      & $(0.40)$      & $(0.50)$      & $(0.41)$      \\
AudienceT3                & $-1.08^{***}$ & $0.21$        & $-1.24^{**}$  & $-0.40$       & $2.94^{***}$  \\
\end{tabular}
\label{tab:RegCode1to5log.int}
\end{center}
\end{table}

\begin{table}[htbp]
\caption{Multilevel Logistic Regression Results for Codes 6 to 10 - Model by Target Linguistic Audience Interaction}
\begin{center}
\begin{tabular}{l c c c c c}
\hline
 & Code 6 & Code 7 & Code 8 & Code 9 & Code 10 \\
\hline
(Intercept)               & $-1.58^{***}$ & $-1.91^{***}$ & $0.89^{***}$  & $-0.75^{***}$ & $-1.83^{***}$ \\
                          & $(0.17)$      & $(0.20)$      & $(0.16)$      & $(0.16)$      & $(0.18)$      \\
ModelM2                   & $0.41$        & $-0.24$       & $-0.00$       & $0.27$        & $-0.06$       \\
                          & $(0.21)$      & $(0.25)$      & $(0.20)$      & $(0.20)$      & $(0.24)$      \\
AudienceT2                & $0.30$        & $1.79^{***}$  & $-1.79^{***}$ & $0.06$        & $-0.37$       \\
                          & $(0.22)$      & $(0.23)$      & $(0.20)$      & $(0.20)$      & $(0.25)$      \\
AudienceT3                & $0.07$        & $0.57^{*}$    & $-0.91^{***}$ & $1.40^{***}$  & $-0.21$       \\
\end{tabular}
\label{tab:RegCode6to10log.int}
\end{center}
\end{table}

\begin{figure}[htbp]
    \centering
    \caption{Interaction Plot for the Interaction of Model and Target Audience on the Likelihood of Flagging Code 4}
    \includegraphics[width = 0.8\linewidth]{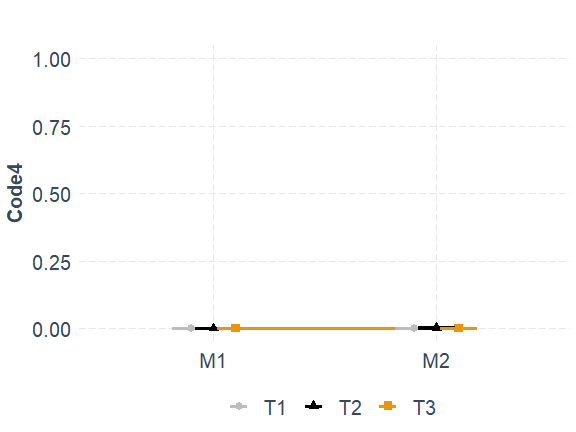}
    \label{fig:Likely4Intplot}
\end{figure}

\begin{table}[htbp]
\caption{Multilevel Logistic Regression Results the Code NOTA (None of the Above) - Model by Target Linguistic Audience Interaction}
\begin{center}
\begin{tabular}{l c}
\hline
 & NOTA \\
\hline
(Intercept)               & $-0.40^{**}$  \\
                          & $(0.13)$      \\
ModelM2                   & $-0.79^{***}$ \\
                          & $(0.19)$      \\
AudienceT2                & $0.95^{***}$  \\
                          & $(0.18)$      \\
AudienceT3                & $0.31$        \\
                          & $(0.18)$      \\
ModelM2:AudienceT2        & $1.04^{***}$  \\
                          & $(0.26)$      \\
ModelM2:AudienceT3        & $0.89^{***}$  \\
                          & $(0.26)$      \\
\hline
AIC                       & $2203.82$     \\
BIC                       & $2241.86$     \\
Log Likelihood            & $-1094.91$    \\
Num. obs.                 & $1692$        \\
Num. groups: Question     & $282$         \\
Var: Question (Intercept) & $0.33$        \\
\hline
\multicolumn{2}{l}{\scriptsize{$^{***}p<0.001$; $^{**}p<0.01$; $^{*}p<0.05$}}
\end{tabular}
\label{tab:RegNOTAlog.int}
\end{center}
\end{table}

\end{appendices}

\end{document}